%% file: article_arxiv.tex
\documentclass[12pt]{article}

\newenvironment{changemargin}[2]{%
\begin{list}{}{%
\setlength{\topsep}{0pt}%
\setlength{\leftmargin}{#1}%
\setlength{\rightmargin}{#2}%
\setlength{\listparindent}{\parindent}%
\setlength{\itemindent}{\parindent}%
\setlength{\parsep}{\parskip}%
}%
\item[]}{\end{list}}

\usepackage{amssymb}
\usepackage{bm}
\usepackage{adjustbox}
\usepackage{multirow}
\usepackage{mathtools}
\usepackage[colorlinks]{hyperref} 
\usepackage{subcaption}
\usepackage{lineno}
\usepackage{setspace}
\usepackage{enumitem}
\usepackage{amsmath}
\usepackage{comment}
\usepackage{bbm}
\usepackage[table]{xcolor}
\usepackage{authblk}

\include{article.definitions}

\usepackage{ulem}
\usepackage{xcolor}

\title{Sparse Semi-supervised Heterogeneous Interbattery Bayesian Analysis}
\author{Carlos Sevilla-Salcedo\thanks{Corresponding author. Email address: sevisal@tsc.uc3m.es}, Vanessa G\'{o}mez-Verdejo, Pablo M. Olmos\footnote{Pablo M. Olmos is also with the Gregorio Mara\~n\'on Health Research Institute.}}
\affil{\small Department of Signal Processing and Communications, Universidad Carlos III de Madrid Legan\'es, 28911 Spain}
\date{\small January 2020}

\begin{document}

\begin{titlepage}
\maketitle

\begin{abstract}
The Bayesian approach to feature extraction, known as factor analysis (FA), has been widely studied in machine learning to obtain a latent representation of the data. An adequate selection of the probabilities and priors of these bayesian models allows the model to better adapt to the data nature (i.e. heterogeneity, sparsity), obtaining a more representative latent space.

The objective of this article is to propose a general FA framework capable of modelling any problem. To do so, we start from the Bayesian Inter-Battery Factor Analysis (BIBFA) model, enhancing it with new functionalities to be able to work with heterogeneous data, include feature selection, and handle missing values as well as semi-supervised problems. 

The performance of the proposed model, Sparse Semi-supervised Heterogeneous Interbattery Bayesian Analysis (SSHIBA) has been tested on 4 different scenarios to evaluate each one of its novelties, showing not only a great versatility and an interpretability gain, but also outperforming most of the state-of-the-art algorithms.
\end{abstract}

\textbf{Keywords:} Bayesian model; Canonical Correlation Analysis; Principal Component Analysis; factor analysis; feature selection; semi-supervised; multi-task

\end{titlepage}

\section{Introduction}
\label{sec:introduction}

Feature Extraction (FE) plays an important role in Machine Learning (ML) with the goal of transforming the original data in a new set with reduced number of features. This is usually carried out by defining a low dimensional latent space where the data is projected. The main advantage of this data transformation relies on the capability of explaining the data information using a significantly lower number of features while removing correlations and noisy components \cite{suk2012novel}. In particular, one method that has been increasingly used in this context is the Canonical Correlation Analysis (CCA) which constructs the latent space from the correlation between two views (two different representations of the data or the data and the target of either a regression or a classification problem). 
Despite commonly used for a single input and output view, its formulation allows to combine the multiple views of the data to improve the extraction of the latent features \cite{zhao2017multi,li2018review,kamronn2015multiview,boutell2004learning, rana2012variational}, what is commonly known as multi-task or multi-view. 

FE algorithms have been adapted to the Bayesian approach where not only the values of all the included variables are obtained, but also their complete distributions are modelled \cite{barendse2014measurement, mezzetti2012bayesian, xue2003bayesian}. This new formulation of FE algorithms known as Factor Analysis (FA) in the Bayesian community, has been widely used in multi-tasks problems such as biomarkers and classification design \cite{pearce2018continuous}, person and digit classification \cite{hernandez2010expectation} or modelling functional neuroimaging data for each subject and then estimate the optimal correlation structure \cite{marquand2014bayesian}. 

Bayesian algorithms provide the additional advantage of facilitating including constraints on the model by defining particular priors over the model variables. For example, the distribution of the latent variables of a FA algorithm can be redefined to impose sparsity on the number of latent factors \cite{li2013exponential, min2018generalized, d2015heart}. This way, the model is capable of automatically determining which latent factors are relevant and eliminate the useless ones. Other approaches can force this to obtain a Feature Selection (FS) so that the model is capable of learning the feature relevance during its training \cite{feng2012bayesian, davis2011bayesian, connor2015biological}. Furthermore, the probabilistic modelling of the data allows to define real data with continuous distributions or categorical data with discrete distributions that are able to capture the nature of the data.
Most methods developed for Bayesian FA centre around working with real data, whereas there is not a wide number of studies about more specific data. In particular, \cite{pauger2019bayesian} presents an algorithm that combines factor analysis with sparsity in the latent space, as well as working with categorical data. By treating the categorical data as whole numbers, the data distribution is able to better fit the original data \cite{terzi2013bayesian}. Conversely, multilabelled methods consider the correlation between labels to model them \cite{gonen2014coupled} reaching certain improvements in the results obtained \cite{gonen2012bayesian, zhang2014augmenting}.

Another considerable advantage of working with the data distribution is that this could be used to impute some values. Semi-supervised learning consist in building the model using data with missing values, this way, on the one hand the model is able to learn from the partial information of this data and, on the other hand, the model estimates these missing values to complete the data. Some algorithms combine this semi-supervised approach with the sparsity in the latent factors \cite{themelis2011novel}. Other methods such as in Gordon et al. \cite{gordon2017bayesian} propose a semi-supervised extension of a Deep Generative Model to obtain a more informative model. Or, both Ge et al. \cite{ge2011semisupervised} and Zhu et al. \cite{zhu2018mixture}, a semi-supervised learning with a Bayesian Principal Component Regression to model soft sensors for industrial applications.

Among the different approaches in the literature for FA and their extensions, the Bayesian Inter-Battery FA model (BIBFA) \cite{klami2013bayesian} has specially attracted our attention since it provides a framework for FA where one can work with multiple data views and sparsity over the latent factors to automatically select the number of latent variables. However, we miss some functionalities in the model to really have a versatile framework able to face any real problem. So, this paper overcomes this limitations proposing a more general formulation able to include the following extensions of the model:
\begin{enumerate}
    \item Endow the model with \textbf{feature selection} capabilities. Our proposal combines the sparsity over the latent space with sparsity over the input feature space by means of a double ARD prior, providing an automatic selection of both latent factors and input features.
    \item Generalise the model data distribution of each view to be able to be adapted to the data nature and work with \textbf{heterogeneous views}. So, the algorithm has also been modified to be capable of working not only with real data but also with multidimensional binary data (Multi-label matrices) and categorical, widening the spectre of problems that can be faced.
    \item A \textbf{semi-supervised} scheme which allows to work with unlabelled data as well as missing data. 
\end{enumerate}
All these proposed extensions of the algorithm can be combined with each other in any way, having a robust formulation of the model, as well as providing an adapted solution for these contexts according to the needs of the problem. This new algorithm with the different extensions is called Sparse Semi-supervised Heterogeneous Inter-battery Bayesian Analysis (SSHIBA). An exemplary notebook, including the complete code of the proposed method, is available at \url{https://github.com/sevisal/SSHIBA.git}.

The article is organised as follows. The BIBFA algorithm of Klami et al. \cite{klami2013bayesian} is reviewed in Section \nameref{sec:related}. A generalised formulation, including all the proposed extensions, is presented in Section \nameref{sec:proposed}. This section just presents the probabilistic model and the inference learning, all mathematical development has been moved to the Appendices. The experimental results, as well as the setup, are presented in Section \nameref{sec:results}, where different databases are used to show the performance of the different versions of the method. Finally, some final remarks and conclusions are given in Section \nameref{sec:conclusion}.

\section{Related Work: Bayesian Inter-Battery Factor Analysis}
\label{sec:related}

In this section we briefly review the Bayesian Inter-Battery Factor Analysis (BIBFA) model, presented in \cite{klami2013bayesian}.

Before introducing the probabilistic formulation of this model, let's present the notation used. For this purpose, given a matrix $\mathbf{A}$ of dimensions $I \times J$, $\mathbf{a}_{i,:}$ represents the $i$-th row of the matrix, $\mathbf{a}_{:,j}$ represents the $j$-th column of the matrix and $a_{i,j}$ represents the $i$-th element of the $j$-th column of the matrix. In case there are different views of the matrix, $\mathbf{A}^{(m)}$ represents the matrix $\mathbf{A}$ of view $m$ and $\mathbf{A}^{\{\mathcal{M}\}}$ represents all the matrices $\mathbf{A}$ of the views in the set $\mathcal{M}$.

\subsection{BIBFA Generative model}
\label{sec:BIBFA}

BIBFA can be understood as a probabilistic CCA in which the effective dimensionality of the projected space is tuned through automatic relevance determination (ARD) priors over the projecting matrices \cite{neal2012bayesian}. Also, as formulated in \cite{klami2013bayesian}, BIBFA handles several observations, each defined as a ``view". The overall goal of CCA is to jointly project all data views into a discriminative low-dimensional space. Assume $\Xnm\in\mathbb{R}^{1 \times D_m}$ is the $m$-th view of the $n$-th data point (each view is a $D_m$-dimensional row vector). If $\mathcal{M}=\{1,2,\ldots,M\}$, then $\XnM = \{\Xnone, \Xntwo,\ldots,\XnMlast\}$ is the complete $n$-th observation. We assume $N$ observations in total. The joint probability density function (pdf) of the BIBFA model is as follows:

\begin{figure}[htp]
  \centering
  \hspace{-1.5cm}
  \begin{subfigure}[t]{0.48\textwidth}
    \centering
    \includegraphics[width=0.9\textwidth]{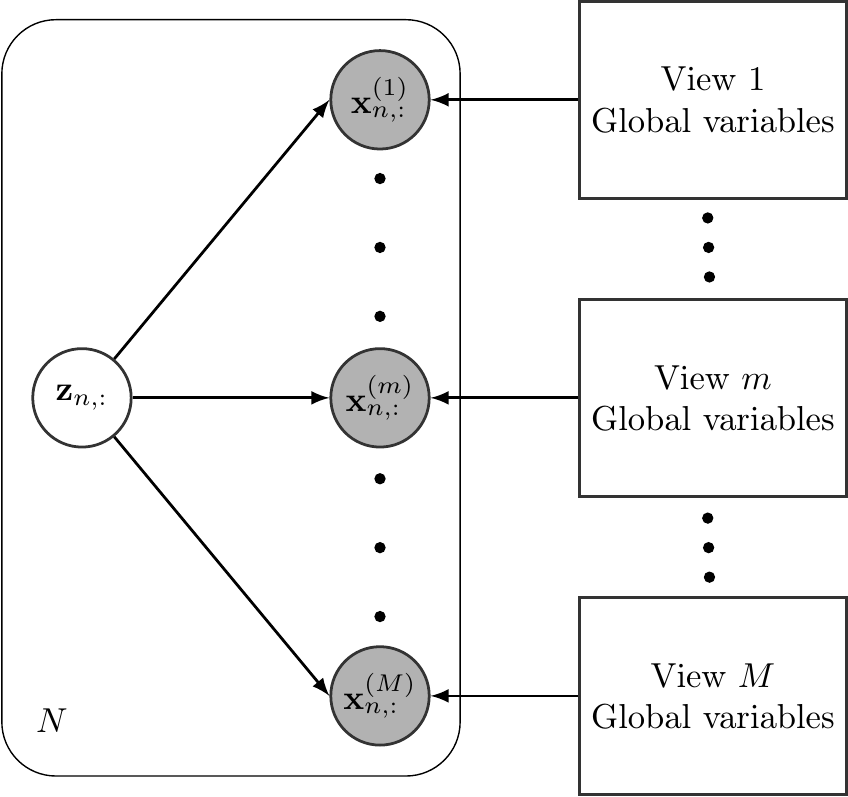}\\
    \caption{Multi-view model.}
    \label{fig:Schemea}
  \end{subfigure}
  ~
  \begin{subfigure}[t]{0.48\textwidth}
    \centering
    \includegraphics[width=1\linewidth]{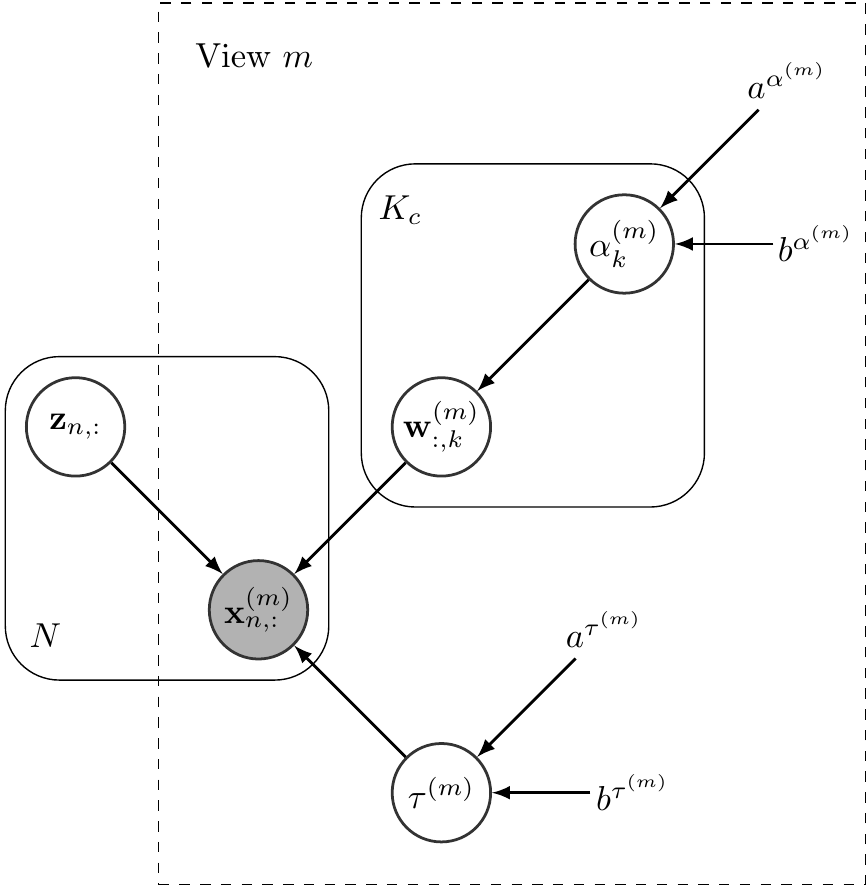}
    \caption{Zoom in view m.}
    \label{fig:Schemeb}
  \end{subfigure}
  \caption{Plate diagram for the BIBFA graphical model. Gray circles denote observed variables, white circles unobserved random variables (rv). The nodes without a circle correspond to the hyperparameters.}
  \label{fig:Scheme}
\end{figure}

\begin{align}
    \Zn \eqsimil \N(0,I_{K_c}) \label{eq:Zprior}\\
	\Wkm \eqsimil \N\p*{0,\p*{\akm}^{-1} I_{K_c}} \label{eq:Wprior}\\
    \Xnm | \Zn \eqsimil \N(\Zn \WmT, \taum^{-1} I_{D_m}) \label{eq:Xprior}\\
    \alpha_{k}^{(m)} \eqsimil \Gamma\p*{a^{\am}, b^{\am}} \label{eq:Alphaprior}\\
	\taum \eqsimil \Gamma\p*{ a^{\taum}, b^{\taum} } \label{eq:Tauprior}
\end{align}
where $I_K$ is an identity matrix of dimension $K$, $\Zn\in\mathbb{R}^{1 \times K_c}$ is the low-dimension latent variable for the $n$-th data point\footnote{Note we work with row-vectors.},  $\Gamma(a,b)$ is a Gamma distribution with parameters $a$ and $b$, $\Wkm$ is the $k$-th column of matrix $\Wm$ (of dimensions $D_m\times K_c$), and up-script $(m)$ corresponds to the $m$-th view. The Gamma distribution over $\alpha_{k}^{(m)}$ enables the model to enforce zero values in order to maximise the model likelihood given our data. Hence, we say that \eqref{eq:Wprior} and \eqref{eq:Alphaprior} form an ARD prior for each of the columns of matrix $\Wm$. The BIBFA graphical model is presented in Figure \ref{fig:Scheme}(a). A closer look on how BIBFA models the generation of each data view is provided in Figure \ref{fig:Scheme}(b). 

In light of the structured found in the posterior distribution of the $\Wm$ matrices, in terms of patterns of columns that are almost all zeros, one can identify common latent factors (elements of $\Zn$) across all views, specific ones only necessary to explain certain views, or irrelevant ones that are not used to explain any view. In \cite{klami2013bayesian}, the latter are removed during inference using a threshold across all views. We will adopt the same strategy, as we later discuss. 

\subsection{BIBFA Variational Inference}

Once the BIBFA generative model is defined, the goal is to evaluate the posterior distribution of all the model variables given the observed data, which is unfeasible due to the intractability of computing the marginal likelihood of the data, i.e. the normalising factor in Baye's rule
\begin{align}\label{post}
p(\Theta|\mathbf{x}_{1,:},\dots, \mathbf{x}_{N,:}) &=\frac{\prod_{n=1}^{N}p(\mathbf{x}_{n,:}|\Theta)p(\Theta)}{p\p*{\mathbf{x}_{1,:},\dots, \mathbf{x}_{N,:}}},\\\label{marglik}
p\p*{\mathbf{x}_{1,:},\dots, \mathbf{x}_{N,:}} &= \int p\p*{\Theta, \mathbf{x}_{1,:},\dots, \mathbf{x}_{N,:}} d(\Theta),
\end{align}
where $\Theta$ comprises all random variables (rv) in the model. In  \cite{klami2013bayesian}, the authors rely on an approximate inference approach through mean-field variational inference \cite{Blei17}, where a lower bound to \eqref{marglik} of the form
\begin{align}\label{elbo}
\log p\p*{\mathbf{x}_{1,:},\dots, \mathbf{x}_{N,:}} \geq \int q(\Theta) \log\left(\frac{\prod_{n=1}^{N}p(\mathbf{x}_{n,:}|\Theta)p(\Theta)}{q(\Theta)}\right)d(\Theta)
\end{align}
is maximised, and a fully factorised variational family is chosen to approximate the posterior distribution in \eqref{post}
\begin{align}
p(\Theta|\XM)
\eqapprox \prodm\p*{q\p*{\Wm} q\p*{\taum} \prodk q\p*{\akm}} \prod_{n=1}^{N}q\p*{\Zn}  \label{eq:qModel}
\end{align}

The mean-field posterior structure along with the lowerbound in \eqref{elbo} results into a feasible coordinate-ascent-like optimization algoritm in which the optimal maximization of  each of the factors in \eqref{eq:qModel} can be computed if the rest remain fixed using the following expression
\begin{align}\label{meanfield}
q^*(\theta_i) \propto \mathbb{E}_{\Theta_{-i}}\left[\log p(\Theta,\mathbf{x}_{1,:},\dots, \mathbf{x}_{N,:})\right],
\end{align}
where $\Theta_{-i}$ comprises all rv but $\theta_i$. This new formulation is in general feasible since it does not require to completely marginalize $\Theta$ from the joint distribution. 

Table \ref{tab:SSHIBA} shows the BIBFA mean-field factor update rules derived in \cite{klami2013bayesian} using \eqref{meanfield}. For a compact notation, we stuck in matrix $\mathbf{Z}$, of dimension $N\times K_c$, the latent projection of all data points and $<>$  represents the mean value of the rv.

\begin{table}[hbt]
\centering
\begin{adjustbox}{max width=\textwidth}
\begin{tabular}{|c|c|c|}
\hline
 & {$\bm{q}^*$ \textbf{distribution}} & \textbf{Parameters} \\\hline
\multirow{3}{*}{$\Zn$}
& \multirow{3}{*}{$\N\p*{\Zn | \mu_{\Zn},\Sigma_{\Z}}$}
& $\mu_{\Zn} = \summ\ang{\taum} \Xm \ang{\Wm} \Sigma_{\Z}$ \\
& & $\Sigma_{\Z}^{-1} = I_{K_c} + \summ\ang{\taum} \ang{\WmT \Wm}$ \\&&\\
\hline                     
\multirow{3}{*}{$\Wm$} 
& \multirow{3}{*}{$\prodd \N \p*{\Wdm | \mu_{\Wdm}, \Sigma_{\Wm}}$} 
& $\mu_{\Wdm} = \ang{\taum} \XmT \ang{\Z} \Sigma_{\Wm}$ \\
& & $\Sigma_{\Wm}^{-1} = \text{diag}(\ang{\am}) + \ang{\taum}\ang{\ZT \Z}$  \\&&\\
\hline    
\multirow{3}{*}{$\akm$}
& \multirow{3}{*}{$\Gamma\p*{\akm | a_{\akm},b_{\akm}}$}
& $a_{\akm} = \frac{D_m}{2} + a^{\am}$ \\ 
& & $b_{\akm} = b^{\am} + \frac{1}{2} \ang{\WmT\Wm}_{k,k}$ \\&&\\
\hline
\multirow{5}{*}{{$\taum$}}
& \multirow{5}{*}{$\Gamma\p*{\taum | a_{\taum},b_{\taum}}$}
& $a_{\taum} = \frac{D_m N}{2} + a^{\taum}$ \\
& & $b_{\taum} = b^{\taum} + \frac{1}{2} \sumn\sumd \Xndm^2 $ \\
& & $ - \Tr\llav*{\ang{\Wm}\ang{\ZT}\Xm}$  \\
& & $ + \frac{1}{2} \Tr\llav*{\ang{\WmT\Wm} \ang{\ZT \Z}}$  \\
\hline
\end{tabular}
\end{adjustbox}
\caption{Updated $q$ distributions for the different rv of the graphical model. These expressions have been obtained using the update rules of the mean field approximation \eqref{meanfield}. See \cite{klami2013bayesian} for further details.
}
\label{tab:SSHIBA}
\end{table}

\subsection{Predictive model}
\label{Sect-BIBFA-predictive}
In addition to only considering real-valued views, the BIBFA model is also limited by the fact that authors do not consider a semi-supervised setting where missing views can be properly handled. To handle missing views, they rely on a training phase, where the posterior distribution of the global variables of the model is computed w.r.t. complete data (i.e. no missing views), to then estimate the distribution of missing views in a test set using a predictive distribution.

Assume use the mean field variational method to approximate the posterior distribution of the BIBFA model parameters $\Theta$ w.r.t. a fully observed training database $\mathcal{D}$, i.e. $q^*(\Theta) \approx p(\Theta|\mathcal{D})$. For a test data point $\mathbf{x}_{\ast,:}$ with observed views contained in the set  $\M_{in}$ and missing views in the set $\M_{out}$, the BIBFA predictive model is as follows. Our first goal is to  
evaluate the marginal posterior probability of the latent projection $\zS$ given $\xminS$
\begin{align} 
p\p*{\zS | \xminS} &= \int p\p{\xmoutS|\zS,\Theta}p\p*{\zS| \xminS,\Theta}p(\Theta|\mathcal{D})d\Theta d\xmoutS\nonumber \\&=\int p\p*{\zS| \xminS,\Theta}p(\Theta|\mathcal{D})d\Theta,
\end{align} 
where note that integration w.r.t. $\xmoutS$ is straightforward as it always integrates to one. Regarding the second term, we can either use Monte Carlo Integration by sampling from $q^*(\Theta)$ or use a point estimate for $\Theta$ (e.g. mean or mode computed from $q^*(\Theta)$). In both cases, once $\Theta$ is fixed, we have that
\begin{align}
    p\p*{\zS| \xminS,\Theta}\propto p\p*{\xminS|\zS,\Theta}p\p*{\zS},
\end{align}
and, since both terms are Gaussian distributions, it is easy to show that $p\p*{\zS| \xminS,\Theta}$ is also Gaussian with mean $\ang{ \zS }$ and covariance matrix $\Sigma_{\zS}$ given by
\begin{align} 
\Sigma_{\zS}^{-1} \eqeq I_{K_c} + \sum_{m\in\M_{in}}\p*{\taum \WmT, \Wm} \\
\ang{ \zS }
\eqeq \sum_{m\in\M_{in}}\p*{\taum \xmS \Wm}\Sigma_{\zS} \nonumber
\end{align} 

We can now write the expression of the distribution of the output views $\xmoutS$ as follows:
\begin{align} 
p\p*{\xmoutS | \xminS,\Theta} \eqeq  \prod_{m\in\M_{out}}p\p*{\xmS | \xminS,\Theta},
\end{align}
where
\begin{align}
p\p*{\xmS | \xminS,\Theta}=\int p\p*{\xmS | \zS,\Theta} p\p*{\zS | \xminS,\Theta} d\zS \label{eq:pXout|Xin}
\end{align} 
where $p\p*{\xmS | \zS,\Theta}$ is defined in \eqref{eq:Xprior}. Using again the properties of the Gaussian distributions we get 
$p\p*{\xmS | \xminS,\Theta}=\N\p*{\xmS  | \mu_{\xmoutS},\Sigma_{\xmoutS}}$, where
\begin{align} 
\Sigma_{\xmoutS}
\eqeq  \taumout^{-1} I_{D_m} + \Wmout \Sigma_{\zS} \WmoutT \label{eq:varXS}\\
\mu_{\xmoutS}
\eqeq  \zS \WmoutT \label{eq:meanXS}
\end{align}

These equations complete the BIBFA standard variational model presented in \cite{klami2013bayesian}, which works in a simple context in which the data matrices are composed of real numbers and can only work on a straight forward manner. The next section is devoted to present our proposal of a generalised BIBFA model able to learn in a semi-supervised fashion, deal with heterogeneous data types and add additional sparsity constraints.

\section{The proposed model: SSHIBA}
\label{sec:proposed}

This section presents the Sparse Semi-supervised Heterogeneous Interbatery Bayesian Analysis (SSHIBA) method. SSHIBA generalises BIBFA in several aspects that we sequentially introduce: 
\begin{enumerate}[label=\arabic*)]
    \item \textbf{Feature selection}:  in addition to being able to automatically select the adequate number of latent variables, by adding a double ARD prior over the matrices $\Wm$, SSHIBA provides automatic relevant determination of both latent factors and input features for each view. 
    \item \textbf{Heterogeneous views}: in contrast to standard BIBFA, which deals observable data views as continuous variables, SSHIBA is able to properly incorporate binary and categorical variables.
    In this way, the model can handle different nature data in the different views.
    \item \textbf{Semi-supervised Learning}: besides, SSHIBA provides the possibility of training the model in a semi-supervised fashion, so that it can properly handle data points with partial observations (some missing views).
\end{enumerate}

These proposed extensions of the method can be combined with each other in any specific way, e.g. combining a multidimensional binary view in which we want to infer some unknown values, as well as doing feature selection. Furthermore, in order to avoid hand-crafted data normalisation, the proposed generative probabilistic model also includes a bias term per view that is learned via variational inference. Namely, in the BIBFA model above, we include the following terms: 
\begin{align}
    \Xnm | \Zn \eqsimil \N(\Zn \WmT + \bim, \taum^{-1} I_{D_m}) \label{eq:Xpriorb}\\
    \bim \eqsimil \N(0,I_{D_m}) \label{eq:bprior}
\end{align}


\subsection{Feature selection in the SSHIBA model}
\label{sec:Sparse}
For this first extension of the method, we propose to redefine the priors of matrix $\Wm$ so that it is able to automatically select both the relevant latent factors and the relevant input features that are used by the model

\subsubsection{Generative model for feature selection}

\begin{figure}[ht]
  \centering
  \includegraphics[page=1,width=0.7\textwidth]{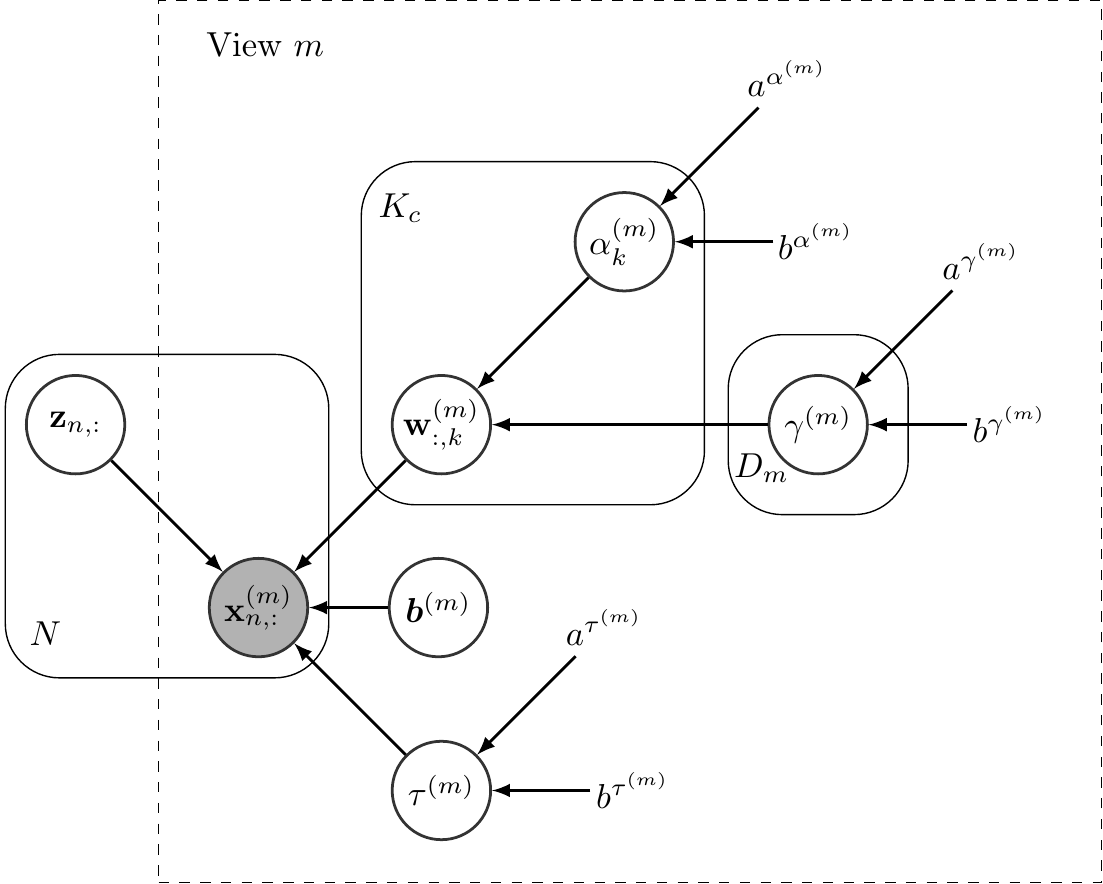}\\
  \caption{SSHIBA's feature selection graphical model.}
  \label{fig:Scheme_sparse}
\end{figure}

To incorporate feature selection capabilities, we propose a double ARD prior over the $\Wm$ matrices, obtaining a different prior over each entry of $\Wm$:

\begin{align}
    \Wdkm  \eqsimil \N\p*{0,\p*{\gamdm\akm}^{-1}} \label{eq:Wprior_sparse} \\
    \gamdm \eqsimil \Gamma\p*{a^{\gamm}, b^{\gamm}} \label{eq:Gammaprior_sparse}
\end{align}
Note that the variance of $\Wdkm$ is the product of two variables: A row-wise prior over $\Wm$, i.e. $\akm$, which was already present in the BIBFA model and is used to perform latent variable selection, and a column-wise prior over $\Wm$, i.e. $\gamdm$ which induces sparsity along the elements of such columns, allowing feature selection interpretability. With the product in \eqref{eq:Wprior_sparse}, we provide the model with the flexibility to find the structural sparsity pattern in $\Wm$ that maximises the evidence. Figure \ref{fig:Scheme_sparse} shows the graphical model of SSHIBA (assuming still real-valued observations).

\subsubsection{Variational inference}

When we augment the BIBFA model presented in Section \ref{sec:BIBFA} with the double ARD method summarized by equations \eqref{eq:Wprior_sparse} and \eqref{eq:Gammaprior_sparse}, we equivalently need to expand accordingly the mean-field posterior distribution, namely
\begin{align}
&p(\Theta|\XM)\eqapprox \nonumber \\
&\prodm\p*{q\p*{\Wm} q\p*{\bim} q\p*{\taum} \prodk q\p*{\akm} \prodd q\p*{\gamdm}} \prod_{n=1}^{N}q\p*{\Zn}.  \label{eq:qModel_sparse}
\end{align}
Table \ref{tab:SSHIBA_s} shows the update rules obtained by applying the mean-field iterative method in \eqref{meanfield} to this new model. We only present those expressions that are now different w.r.t. the mean-field expressions for the BIBFA model. A detail calculation of these expressions can be found in Appendix A (available as supplementary material).

\begin{table}[hbt]
\centering
\begin{adjustbox}{max width=\textwidth}
\begin{tabular}{|c|c|c|}
\hline
& {$\bm{q}^*$ \textbf{distribution}} & \textbf{Parameters} \\\hline
\multirow{3}{*}{$\Zn$}
& \multirow{3}{*}{$\N\p*{\Zn | \mu_{\Zn},\Sigma_{\Z}}$}
& $\mu_{\Zn} = \summ\p*{\ang{\taum}\p*{\Xm - \mathbbm{1}_N\ang{\bim}} \ang{\Wm}}\Sigma_{\Z}$ \\
& & $\Sigma_{\Z}^{-1} = I_{K_c} + \summ\ang{\taum} \ang{\WmT \Wm}$ \\
\hline  
{\multirow{3}{*}{$\Wm$}}
& \multirow{3}{*}{$\prodd\N \p*{\Wdm | \mu_{\Wdm}, \Sigma_{W_d^{(m)}}}$}
& $\mu_{\Wm} = \ang{\taum} \p*{\Xm - \mathbbm{1}_N \ang{\bim}}^T \ang{\Z} \Sigma_{\Wm}$ \\
& & $\Sigma_{W_d^{(m)}}^{-1} = \text{diag}(\ang{\am})\ang{\gamdm} + \ang{\taum}\ang{\ZT \Z}$ \\
\hline
\multirow{3}{*}{$\bim$}
& \multirow{3}{*}{$\N \p*{\bim | \mu_{\bim}, \Sigma_{\bim}}$}
& $\mu_{\bim} = \ang{\taum} \sumn\p*{\Xnm - \ang{\Zn}\ang{\WmT}} \Sigma_{\bim} $ \\
& & $\Sigma_{\bim}^{-1} = \p*{N \ang{\taum} + 1} I_{D_m}$ \\
\hline
\multirow{3}{*}{$\am$}
& \multirow{3}{*}{$\prodk \Gamma\p*{\akm | a_{\akm},b_{\akm}}$}
& $a_{\akm} = \frac{D_m}{2} + a^{\am}$ \\
& & $b_{\akm} = b^{\am} + \frac{1}{2} \sumd\ang{\gamdm}\ang{\Wdkm\Wdkm}$ \\
\hline
\multirow{6}{*}{{$\taum$}}
& \multirow{6}{*}{$\Gamma\p*{\taum | a_{\taum},b_{\taum}}$}
& $a_{\taum} = \frac{D_m N}{2} + a^{\taum}$ \\
& & $b_{\taum} = b^{\taum} + \frac{1}{2} \sumn\sumd \Xndm^2 $ \\
& & $ - \Tr\llav*{\ang{\Wm}\ang{\ZT}\Xm} + \frac{1}{2} \Tr\llav*{\ang{\WmT\Wm} \ang{\ZT \Z}}$  \\
& & $ - \sumn \Xnm \ang{\bimT} + \sumn \ang{\Zn} \ang{\WmT} \ang{\bimT}+ \frac{N}{2} \ang{\bim \bimT}  $  \\
\hline
\multirow{3}{*}{$\gamm$}
& \multirow{3}{*}{$\prodk \Gamma\p*{\gamdm | a_{\gamdm},b_{\gamdm}}$}
& $a_{\gamdm} = \frac{K_c}{2} + a^{\gamm}$ \\
& & $b_{\gamdm} = b^{\gamm} + \frac{1}{2} \sumk\ang{\akm}\ang{\Wdkm\Wdkm}$ \\
\hline
\end{tabular}
\end{adjustbox}
\caption{Distribution $q$ of the different rv of the graphical model for feature selection together with the different distribution parameters. Where $\mathbbm{1}_N$ is a row vector of ones of dimension $N$.}
\label{tab:SSHIBA_s}
\end{table}

Since variable $\gamm$ provides a measure of importance for each feature (higher $\gamm$, lower importance), the  model is now capable of providing a measure of the relevance of each feature. In other words, this version allows the model to provide an online feature ranking or feature selection for any input data, improving the interpretability of the results. 

Finally, note that, given the modified predictive distribution $q^*(\Theta)$ in  Table \ref{tab:SSHIBA_s}, the predictive model remains the same w.r.t. the BIBFA predictive model in Section \ref{Sect-BIBFA-predictive}.


\subsection{Heterogeneous data: Multidimensional binary views}

In this section we propose an extension of the model that is capable of modelling any of the data views as a multidimensional binary rv. For example, this extension can be used to model the output view of a multi-label classification problem. 

\subsubsection{Generative model}

\begin{figure}[ht]
  \centering
  \includegraphics[page=1,width=0.6\textwidth]{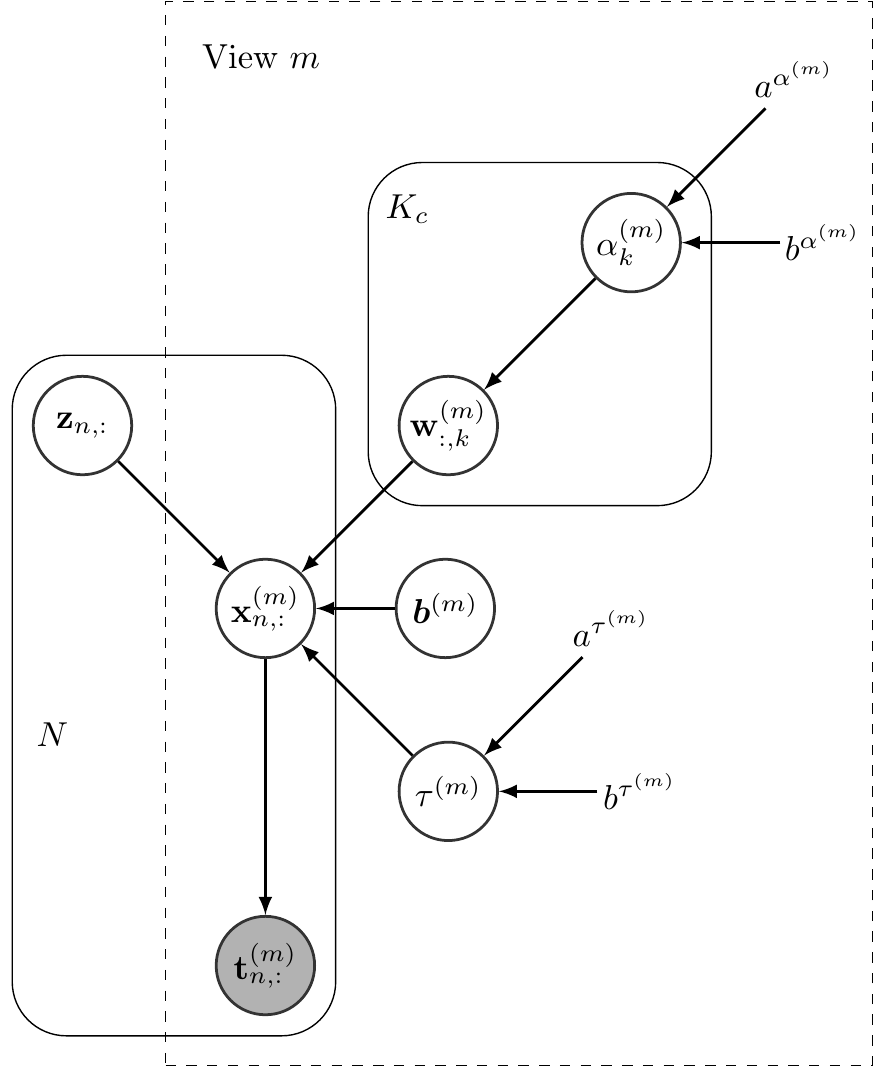}\\
  \caption{SSHIBA graphical model for multi-dimensional binary views.}
  \label{fig:SchemeMultilabel}
\end{figure}

To accommodate the model for binary views, we incorporate the Bayesian logistic regression model presented in \cite{jaakkola1997variational}, as it is summarised in the graphical model of Figure \ref{fig:SchemeMultilabel}. Observe now that variable $\Xnm$ is now unobserved but still keeps the same conditional distribution \eqref{eq:Xpriorb}; i.e. $\Xnm$ is still a $D_m$-real valued vector following a Gaussian distribution given $\Zn$.
Furthermore, we introduce a new observed variable binary vector $\tnm$, also of dimension $D_m$, whose conditional distribution given $\Xnm$ is a product of logistic regression terms
\begin{align}
    p\p*{\tnm|\Xnm} & = \prod_{d=1}^{D_m} p\p*{\tndm|\Xndm}\\ \label{eq:binviews}
    p\p*{\tndm|\Xndm} & = \sigma\p*{\Xndm}^{\tndm}\p*{1-\Xndm}^{1-\tndm} = e^{\Xndm \tndm} \sigma\p*{-\Xndm},
\end{align}
where $\sigma\p*{a} = (1+e^{-a})^{-1}$. Following \cite{jaakkola1997variational}, to develop the variational machinery for the observation model in \eqref{eq:binviews}, we will use the following lower bound on the logistic regression conditional probability
\begin{align}
p\p*{\tndm|\Xndm}
\eqeq e^{\Xndm \tndm} \sigma\p*{-\Xndm} \geq \nonumber \\
\eqline e^{\Xndm \tndm} \sigma\p*{\xi_{n,d}} e^{-\frac{\Xndm + \xi^{(m)}_{n,d}}{2}  - \lambda \p*{\xi^{(m)}_{n,d}} \p*{\xndmtwo - \xi_{n,d}^2}} \label{eq:ptnd}
\end{align}
where $\lambda \p*{a} = \frac{1}{2a}\p*{\sigma\p*{a} - \frac{1}{2}}$, and $\xi^{(m)}_{n,d}$ are variational parameters that are optimized by maximizing the lower bound in \eqref{elbo} as shown in Appendix B. Using this bound, we can lower bound $ p\p*{\tm|\Xm}$ as follows
 \begin{align}
 &p\p*{\tm|\Xm} \geq h\p*{\Xm,\bm{\xi}} = \nonumber\\
 &\prod_{n=1}^N\prod_{d=1}^{D_m} \p*{\sigma\p*{\xi_{n,d}} e^{\Xndm t_{nd}^{(m)} -\frac{\Xndm + \xi_{n,d}}{2}  - \lambda \p*{\xi_{n,d}} \p*{X_{nd}^{(m)^2} - \xi_{n,d}^2}}}. \label{eq:pt}
 \end{align}

\subsubsection{Variational inference}

Given the graphical model in Fig. \ref{fig:SchemeMultilabel}, the mean-field variational family that is used is as follows
\begin{align}
&p\p*{\Theta | \tmt,\mathbf {X}^{\{\mathcal{M}_r\}}} \eqapprox \nonumber \\
& q\p*{\Z} \prod_{m_t \in \mathcal{M}_t} \left(\prod_{n=1}^{N}q\p*{\xnmmt}\right) \prodm q\p*{\Wm} q\p*{\bim} q\p*{\am} q\p*{\taum} q\p*{\gamm}, \label{eq:qModel_mul}
\end{align}
where $\mathcal{M}_t$  is the set of views in which we want to have the multidimensional binary data and $\mathcal{M}_r$ are the rest of the views. The details about the mean-field variational updates can be found in Appendix B. Note that, condition to a fixed $\Xmmt$, the model is equivalent to the case of real-valued observations and, hence, most of the mean-field updates remain almost the same, as long as we replace in Table \ref{tab:SSHIBA} and \ref{tab:SSHIBA_s} $\xnmmt$ (or the stacked data matrix $\XmT$) by the mean $\ang{\xnmmt}$ ($\ang{\XmT}$) determined by the current $q\p*{\xnmmt}$ distribution for each data point. Regarding this latter term, the variational update-rule is given in Table\ref{tab:SSHIBA_multilabel}.
\begin{table}[hbt]
\centering
\begin{adjustbox}{max width=\textwidth}
\begin{tabular}{|c|c|c|}
\hline
& {$\mathbf {q}$ \textbf{distribution}} & \textbf{Parameters} \\\hline
\multirow{2}{*}{$\xnmmt$}
& \multirow{2}{*}{$\N \p*{\xnmmt | \mu_{\xnmmt},\Sigma_{\Xmmt}}$ }
& $\mu_{\xnmmt} = \p*{\tndotmmt - \frac{1}{2} + \ang{\taummt}\ang{\Zn}\ang{\WmmtT} + \ang{\bim}}\Sigma_{\xnmmt}$ \\
& & $\Sigma_{\Xmmt}^{-1} = \ang{\taummt} I + 2\Lambda_{\text{\boldmath$\xi$}_{n,:}}$ \\
\hline
\end{tabular}
\end{adjustbox}
\caption{Mean-field update rule for the $q\p*{\xnmmt}$ distribution in \eqref{eq:qModel_mul}, where $\Lambda_{\bm{\xi}_{n,:}}$ is a diagonal matrix for which the diagonal elements are $\lambda\p*{\xi_{n,1}}, \lambda\p*{\xi_{n,2}}, \dots, \lambda\p*{\xi_{n,D_m}}$. This distribution only affects the views modelled as multidimensional binary data.}
\label{tab:SSHIBA_multilabel}
\end{table}

Unlike the BIBFA predictive distribution in Section\ref{Sect-BIBFA-predictive}, SSHIBA with multi-dimensional binary observation requires approximate inference (e.g. variational inference or Monte Carlo) to estimate the posterior latent distribution w.r.t. to the observe data. This case can be directly reformulated from the semi-supervised SSHIBA model presented in Section \ref{Sect-semisupervised}, and hence we omit it here to avoid uncluttered notation.


\subsection{Heterogeneous data: Categorical observations}

In this section we present a model that is capable of working with categorical data, in any of the data views.

\subsubsection{Generative model}

\begin{figure}[ht]
  \centering
  \includegraphics[page=1,width=0.6\textwidth]{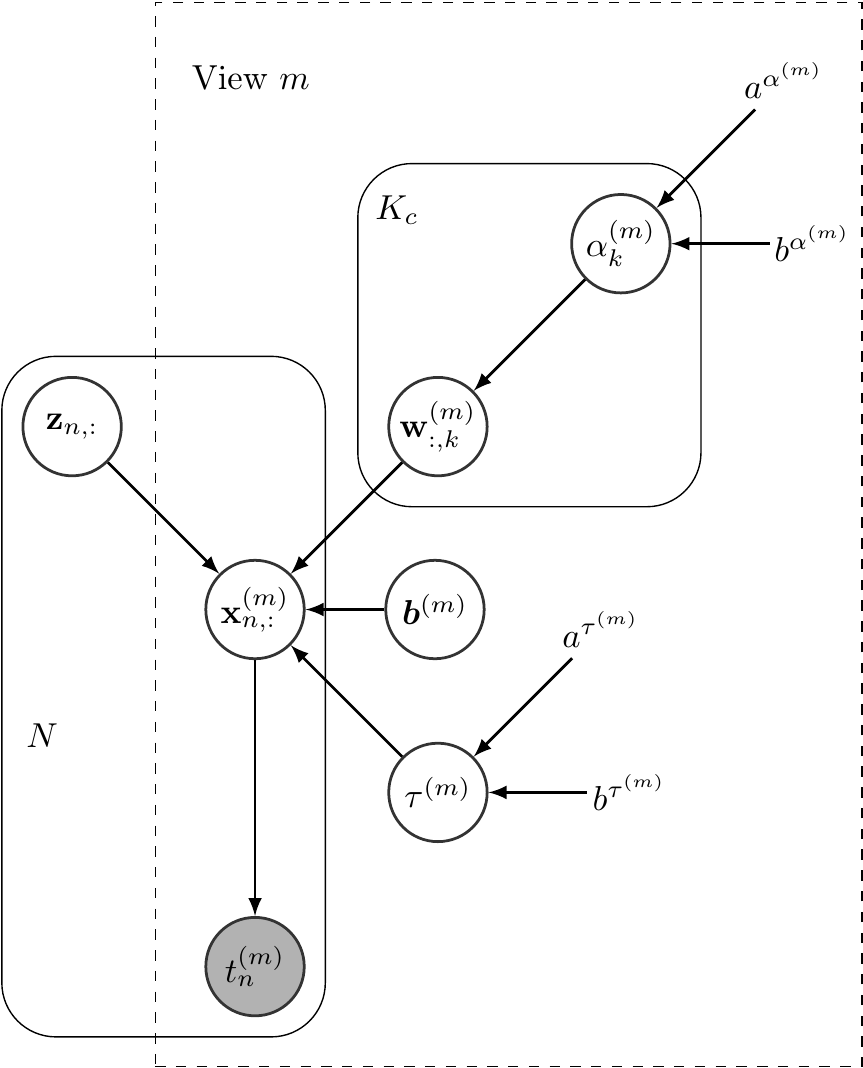}\\
  \caption{SSHIBA graphical model for categorical views.}
  \label{fig:SchemeCategorical}
\end{figure}

We incorporate the multinomial probit in \cite{girolami2006variational} to accommodate our model for categorical observations. In this case, the structure is similar to the one followed by the multidimensional binary case of Figure \ref{fig:SchemeMultilabel} but, in the categorical case, $\ttnm$ (assuming that the $m$-th view corresponds to a categorical variable) is an integer scalar that takes values in the set $\{0,\dots, D_m -1\}$, being $D_m$ the number of classes. The multinomial probit relates $\Xnm$ with $\ttnm$ as follows:
\begin{align}
\ttnm = i \hspace{1cm} if \hspace{1cm} \Xnim = \max_{1\leq d \leq D_m}\p*{\Xndm}.
\end{align}

If we set the noise parameter $\taum = 1$,  in \cite{girolami2006variational} it is shown that we can express $p\p*{\ttnm=i | \Zn, \Wm}$ as follows:
\begin{align}
p\p*{\ttnm=i | \Zn, \Wm} = \E_{p\p*{u}}\cor*{\prod_{j\neq i}\p*{\Phi\p*{u + \ynim - \ynjm}}}
\end{align}
where $\ynm = \Zn \WmT$, $p\p*{u} \sim \N\p*{0,1}$, and $\Phi(\cdot)$ is the standard Gaussian cumulative distribution function (cdf). Expectations w.r.t. $p(u)$ can be effectively approximated using Monte Carlo, as they only require sampling from a uni-dimensional standard Gaussian.

\subsubsection{Variational inference}

Deriving mean-field update for the categorical views closely follows the methodology in \cite{girolami2006variational}, so we omit further details here. Given the  mean-field variational family in \eqref{eq:qModel_mul} (assuming now that $\mathcal{M}_t$ is the set of views that correspond to categorical observations), the mean-field update of the term $q\p*{\xnmmt}$ is summarized in Table \ref{tab:SSHIBA_multilabel}. The mean-field update for the rest of the terms are provided in previous sections (as in the multi-dimensional binary case, we replace $\XmT$ by $\ang{\XmT}$). Observe that, given $\ttnm$, $q\p*{\xnmmt}$ corresponds to a truncated Gaussian distribution.

Again, we note that a predictive model will be easily formulated from the semi-supervised case presented in the next subsection.

\begin{table}[hbt]
\centering
\begin{adjustbox}{max width=\textwidth}
\begin{tabular}{|c|c|c|}
\hline
& {$\mathbf {q}$ \textbf{distribution}} & \textbf{Moments} \\\hline
\multirow{3}{*}{$\xnmmt$}
& $\frac{1}{\bm{\xi}_{n,:}} \N \p*{\xnmmt | \ang{\ynmt},I}\times$
& $\ang{\Xnimt} = \ang{\ynimt} + \sum_{j\neq i}\p*{\ang{\ynjmt}-\ang{\Xnjmt}}$ \\
& \multirow{2}{*}{$\delta\p*{\Xnimt > \Xnjmt \forall i\neq j}$} 
& $\ang{\Xnjmt} = \ang{\ynjmt} - \frac{1}{\bm{\xi}_{n,:}}\E_{p\p*{u}}\bigl[\N_u\p*{\ang{\ynjmt}-\ang{\ynimt},1}\bigr.$ \\
& & \hspace{1cm} $\bigl.\prod_{k\neq i \neq j}\p*{\Phi\p*{u+\ang{\ynimt}-\ang{\ynkmt}}}\bigr]$ \\
\hline
\end{tabular}
\end{adjustbox}
\caption{$q$ distribution of the different rv of the graphical model for the categorical scheme, where $\ang{\ynmt} = \ang{\Zn} \ang{\WmT} + \ang{\bim}$ and $\bm{\xi}_{n,:} = \E_{p\p*{u}}\cor*{\prod_{j \neq i}\p*{\Phi\p*{u+\ang{\ynimt}-\ang{\ynjmt}}}}$ and assuming that $\ttnm = i$. This distribution only affects the views modelled as categorical data.}
\label{tab:SSHIBA_Categorical}
\end{table}


\subsection{Semi-supervised SSHIBA}
\label{Sect-semisupervised}

The last main contribution of the paper is to show how missing-views can be incorporated into SSHIBA training (e.g. variational inference) following an unsupervised fashion, in which there is no need for a predictive distribution since both ``training" and ``test" data are jointly fused by the model, which simply considers as unobserved both the views in the test data that we aim at predicting and the missing values in both ``training" and ``test" sets. Semi-supervised SSHIBA  can also handle feature selection and both real, binary and categorical views.

In the case the m-th view corresponds to a real-variable, we denote by $\XmS$ (in contrast to $\Xm$) to the set of data points for which this view is missing. Similarly, if the m-th corresponds to a multi-dimensional binary variable or categorical variable, the set of data points for which this view is missing is denoted by $\tmS$ (in contrast to $\tm$). Note that the SSHIBA graphical model summarized in Figures \ref{fig:Scheme_sparse}, \ref{fig:SchemeMultilabel}, and \ref{fig:SchemeCategorical} remains unaltered, we simply have white dots instead of grey dots for those data points for which the corresponding view is unobserved.

\subsubsection{Variational inference}

Missing views are handled as any other random variable in the model and hence during variational inference our goal is now to approximate the joint posterior distribution of the parameters of the model $\Theta$ and the missing data views ($\XmS$ or  $\tmS$). Following the mean-field method, we again assume a variational family that factorizes accross all elements in $\Theta$ and all data points in $\XmS$ or $\tmS$: 

\begin{align}
p(\Theta, \mathbf {\tilde{T}}^{\{\mathcal{M}_t\}},\mathbf {\tilde{X}}^{\{\mathcal{M}_r\}} &| \tmt,\mathbf {X}^{\{\mathcal{M}_r\}}) \eqapprox \nonumber \\
& q\p*{\Z} \prod_{m_t \in \mathcal{M}_t} \left(q(\mathbf {\tilde{T}}^{(m_t)})\prod_{n=1}^{N}q\p*{\xnmmt}\right) \nonumber\\
&\times \prod_{m_r \in \mathcal{M}_r}q(\mathbf {\tilde{X}}^{(m_r)})\prodm\p*{q\p*{\Wm} q\p*{\am} q\p*{\taum} q\p*{\gamm}} \label{eq:qModel_SS}. 
\end{align}

The mean-field update for the different terms can be found in Appendix C and the final distributions are shown in in Table \ref{tab:SSHIBA_SS}.
\begin{table}[hbt]
\centering
\begin{adjustbox}{max width=\textwidth}
\begin{tabular}{|c|c|c|c|}
\hline
\textbf{Version}& \textbf{Variable} & { $\mathbf {q}$ \textbf{distribution}} & \textbf{Parameters} \\\hline
\multirow{2}{*}{\textit{Regression}} &
\multirow{2}{*}{\large$\XmS$}
& \multirow{2}{*}{$\prodn \N\p*{\XnmS | \mu_{\XnmS}, \Sigma_{\XmS}}$}
& $\mu_{\XmS} = \ang{ \ZS } \ang{ \Wm }^T$ \\
& & & $\Sigma_{\XmS} = \ang{ \taum }^{-1} I_{D_m}$ \\
\hline
\multirow{3}{*}{\textit{Multidimensional}} &
\multirow{3}{*}{\large$\tmS$}
& \multirow{3}{*}{$\prodn \N \p*{\tnmS | \ang{\tnmS},\Sigma_{\tmS}}$}
& $\mu_{\tnmS} = \sigma\p*{\ang{\XmS}}$ \\
& & & $\Sigma_{\tmS} = \frac{e^{\ang{\XmS}}}{\p*{1+e^{\ang{\XmS}}}^2}$ \\
\hline
\multirow{2}{*}{\textit{Categorical}} &
\multirow{2}{*}{\large$\ttmS$}
& \multirow{2}{*}{$\prodn \N \p*{\ttnmS | \ang{\ttnmS},\Sigma_{\ttmS}}$} & $\ang{\ttnmS} = \ang{\ySnjmt} - \frac{1}{\bm{\xi}_{n,:}}\E_{p\p*{u}}\bigl[\N_u\p*{\ang{\ySnjmt}-\ang{\ySnimt},1}\bigr.$ \\
& & & \hspace{1cm} $\bigl.\prod_{k\neq i \neq j}\p*{\Phi\p*{u+\ang{\ySnimt}-\ang{\ySnkmt}}}\bigr]$ \\
\hline
\end{tabular}
\end{adjustbox}
\caption{$q$ distribution of the different rv of the graphical model for the semi-supervised scheme. The table shows what are the different parameters of the distributions. The first parameter is the mean and the second one is the variance. Where $\ang{\ySnmt} = \ang{\ZnS} \ang{\WmT} + \ang{\bim}$.}
\label{tab:SSHIBA_SS}
\end{table}

\section{Results}
\label{sec:results}

In this section we present the experimental results that demonstrate the ability of SSHIBA to capture the statistical properties of real databases, while comparing it with some state-of-the-art algorithms. We divide our experiments in four different scenarios, in which we focus on different aspects of the model.
\begin{itemize}
\item SSHIBA for multilabel/categorical prediction: Comparison of several versions of the algorithm with different reference methods or baselines over datasets from different nature. We  also investigate the performance when the amount of available data is small. 
\item Feature selection with SSHIBA: We use a face recognition dataset to learn which features are the most relevant.
\item Missing data imputation with SSHIBA: We study the ability of the proposed method to impute missing data in a real dataset. 
\item Multiview learning with SSHIBA: In this case we study the benefit of treating each data view independently, compare to the case when multiple views are join together.
\end{itemize}

First, we define a measure to compare with the baselines. As, in general, we work with multiclass datasets, we decided to use the balanced Multiclass Area Under the Curve (AUC) calculated as $AUC_{mc} = \frac{1}{N} \sum_c\p*{N_c \times AUC_c}$, where $N$ is the total number of samples, $N_c$ is the number of samples of class $c$ and $AUC_c$ is the AUC of class $c$ with respect to the rest of the classes.

To do so, we first defined a measure to compare with the baselines. As, in general, we work with multiclass datasets, we decided to use the balanced Multiclass Area Under the Curve (AUC) calculated as $AUC_{mc} = \frac{1}{N} \sum_c\p*{N_c \times AUC_c}$, where $N$ is the total number of samples, $N_c$ is the number of samples of class $c$ and $AUC_c$ is the AUC of class $c$ with respect to the rest of the classes.

We implemented the SSHIBA algorithm so that it can make automatic latent factor selection, also referred as pruning. For this purpose, during the inference learning we remove the $k$ columns of $\Wm$,  $\forall m$, if all the elements of $\Wkm$, across all the views, are lower than the pruning threshold. For our experiments, this pruning threshold was set to $10^{-6}$.

To determine the number of iterations of the inference process, we used a convergence criteria based on the evolution of the lower bound. In particular, we stop the algorithm either when $LB[-2] > LB[-1](1 - 10^{-8})$, where $LB[-1]$ is the lower bound at the last iteration and $LB[-2]$ at the previous one, or when it reaches $5*10^{4}$ iterations.

Both the SSHIBA and BIBFA algorithms were trained 10 times with different initialisations, keeping the one with the best lower bound. 

The implementation of this project was done using \textit{Python 3.7} and the different baselines as well as train and test splits where carried out using packages from \textit{Scikit-learn} \cite{scikit-learn}. 

\subsection{Database description}
As the presented model works in a wide range of contexts, we included several databases of different nature (different sizes, dimensions, types of variables, ...) to check its performance over a a wide range of scenarios. 

First of all, we used two databases from the Mulan repository \cite{mulan}: the \textit{yeast} database \cite{elisseeff2002kernel} and the \textit{scene} database \cite{boutell2004learning}. These are multilabel problems with numeric features, so they work with heterogenous views. We also worked with the a categorical \textit{AVIRIS} database \cite{baumgardner220} composed of 220 Band Hyperspectral Image of agronome farms.

For some scenarios, we used the Labeled Faces in the Wild (LFW) dataset \cite{LFWTech}. The data, in this case, is composed of face photographs of different people. We used an aligned version of the dataset obtained by \cite{wolf2010effective} in order to work with images that are comparable. At the same time, the images have been cropped to eliminate undesirable information and resized to reduce the computational cost of training the models, having images of $60 \times 40$ pixels. Once the images were processed we decided to work with two different problems:
\renewcommand{\labelitemi}{$\diamond$}
\begin{itemize}
    \item \textbf{Face recognition}: It consists in identifying the person to whom the image corresponds. In this case we used the 7 people having most images in the dataset and the labels for the images is the person who is the photo. We will refer to this version as LFW.
    \item \textbf{Multilabel attributes}: It consists in determining whether an image has certain attributes or not. The attributes, obtained by \cite{kumar2009attribute}, correspond to different physical information related to the people in the photographs, such as gender, hair colour or wearing glasses. Therefore, we have a set of binary labels corresponding to the different attributes of the image. We will refer to this version as LFWA.
\end{itemize}
The characteristics of these databases are also described in Table \ref{tab:database}.

\begin{table}[hbt]
\centering
\begin{adjustbox}{max width=\textwidth}
\begin{tabular}{|r||ccc|}
\hline
Database & Samples & Features & Labels  \\\hline\hline
\textit{yeast}  & 2,417 & 103 & 14 \\
\hline    
\textit{scene}  & 2,407 & 294 & 6 \\
\hline  
\textit{AVIRIS}  & 21,025 & 220 & 16 \\
\hline  
\textit{LFW}    & 1,277 & 2.400 & 7 \\
\hline  
\textit{LFW-A}  & 22,343 & 2.400 & 73 \\
\hline  
\end{tabular}
\end{adjustbox}
\caption{Summary of the main characteristic of the databases used in this work.}
\label{tab:database}
\end{table}
The performance of the method has been measured using train and test sets. In particular, both the \textit{scene} and \textit{yeast} databases are already divided into train and test sets, around a $50\%$ and $60\%$ train data respectively. In the case of the \textit{LFW} databases as well as \textit{AVIRIS}, they were split using $70 \%$ train / $30 \%$ test partitions. At the same time, a 10 folds Cross-Validation (CV) was used to adjust the regularization parameter for the logistic regression, MLP and ridge regression. The number of latent factors of the PCA has been set to those who explain $95\%$ of the variance.

For one of the experiments carried out the training dataset is subsampled to prove the viability of the method with low density data. To do so, we use the iterative stratifier presented in \cite{sechidis2011stratification} to have splits with the minority categories properly represented.

\subsection{Baseline or state-of-art methods}
\label{Sect-baselines}
To analyse the different versions of the method in comparison to some contextual results, we decided to include some state-of-the-art algorithms to obtain reference scoring. In particular, we have used the following methods:
\begin{itemize}
    \item \textbf{CCA}: The Canonical Correlation Analysis (CCA) is a supervised feature extraction method which finds a latent space for the data. Due to the parallelisms with our method,  we decided to used this algorithm as one of the baselines to compare to.
    \item \textbf{Linear ridge regression}: As the BIBFA method is based on a linear estimation, we decided to compare our results with a linear ridge regression as a classifier (linear ridge regression + a cross-validated threshold).
    \item \textbf{Logistic regression}: As all of the problems to solve involve classification tasks, we have include this state-of-the-art method widely used as a classifier.
    \item \textbf{MLP}: To compare our results to those of a neural network, we used a Multi-Layer Perceptron (MLP) with one layer.
    \item \textbf{BIBFA}: We also included the base method presented in \cite{klami2013bayesian}. As they indicate in the paper, we added a final thresholding process to obtain a label prediction.
\end{itemize}

Furthermore, for the data imputation section we decided to compare our results to three standard imputation approaches: substituting by the mean, the median or the most frequent value.

\subsection{SSHIBA for multilabel/categorical prediction}

In this experiment we used both the \textit{yeast} multilabel database and the \textit{AVIRIS} categorical database to test the algorithms in different scenarios. The initial number of latent factors for both SSHIBA and BIBFA was set to 100 and 200 respectively. For both databases we used the real view to predict either the multilabel or the categorical data. First of all, we compared our multilabel approach (SSHIBA) to all baseline methods introduced in Section \ref{Sect-baselines}. For both the SSHIBA and BIBFA, we perform test estimation using the the standard predictive approach described in Section \ref{Sect-BIBFA-predictive}.  Furthermore, all these results have been calculated using the complete dataset, as well as a reduced version consisting of a $20\%$ of the original data.

\begin{table}[thp]
\centering
\begin{adjustbox}{max width=\textwidth}
\begin{tabular}{c|c|c}
\hline
 & Complete dataset & $20\%$ of data  \\\hline \hline
\multirow{2}{*}{\textit{SSHIBA}} & 0.66 & $\mathbf{0.65 \pm 0.005}$\\
 & \cellcolor{gray!10} 66 & \cellcolor{gray!10} $20 \pm 2$\\ \hline
\multirow{2}{*}{\textit{BIBFA}} & \textbf{0.69} & $0.63 \pm 0.008$\\
 & \cellcolor{gray!10} 66 & \cellcolor{gray!10} $29 \pm 1$\\
\hline     \hline
\multirow{2}{*}{CCA}& 0.61 & $0.56 \pm 0.008$ \\ 
 & \cellcolor{gray!10} 13 &  \cellcolor{gray!10} 13 \\ \hline
\multirow{2}{*}{CCA + Log. Reg.}& 0.66 & $0.60 \pm 0.012$\\ 
&\cellcolor{gray!10}  13 & \cellcolor{gray!10} 13 \\ \hline
\multirow{2}{*}{PCA + Log. Reg.} & 0.68 & $\mathbf{0.65 \pm 0.005}$ \\ 
 & \cellcolor{gray!10} 73 & \cellcolor{gray!10} $66 \pm 1$\\ \hline
\multirow{2}{*}{MLP} & 0.61 & $0.59 \pm 0.005$ \\ 
 & \cellcolor{gray!10} 300 & \cellcolor{gray!10} $220 \pm 98$\\ \hline
Logistic reg. & 0.67 & $\mathbf{0.65 \pm 0.005}$\\ \hline 
Ridge reg. & 0.68 & $\mathbf{0.65 \pm 0.006}$\\ \hline 
\hline
\end{tabular}
\end{adjustbox}
\caption{Results on \textit{yeast} database of the predictive SSHIBA and the different methods under study. Results include the performance in terms of AUC (white cells) and the number of latent factors (grey cells) when the complete dataset is used and when only $20\%$ of the training samples are used. The results on the reduced dataset have been calculated 5-fold CV, so their standard deviations are also included.}
\label{tab:yeastTotal}
\end{table}

In Table \ref{tab:yeastTotal} we can see the results obtained with the \textit{yeast} database. We include both the performance and the dimensionality of the latent space, automatically tuned by either SSHIBA or BIBFA. The results on the complete dataset (left column) provide an insight on the method, where we can see that the algorithm is capable of providing a dimensionality reduction of the input features while maintaining the prediction performance compared to the rest of the discriminative approaches. 
Furthermore, we can see that the bayesian approach provided by SSHIBA makes it more robust when we consider a smaller training set (right column), providing the exact same prediction performance than PCA+Logistic regression, Logistic regression and Ridge regression, with a latent dimensionality significantly smaller than PCA + Logistic regression (20 vs 66). We conjecture that the ability of SSHIBA to treat each data type according to its true nature (binary/categorical) explains the robustness of the method in the low sample-size regime (good performance + small latent dimensionality).

\begin{table}[thp]
\centering
\begin{adjustbox}{max width=\textwidth}
\begin{tabular}{c|cc}
\hline
 &  Complete dataset & $20\%$ of data \\\hline \hline
\multirow{2}{*}{\textit{SSHIBA - multilabel}} & 0.88 & $0.85 \pm 0.014$ \\
 &\cellcolor{gray!10}  194 &\cellcolor{gray!10} $189 \pm 21$ \\ \hline
\multirow{2}{*}{\textit{SSHIBA - categorical}} & \textbf{0.89} & $0.87 \pm 0.007$ \\
 & \cellcolor{gray!10} 197 & \cellcolor{gray!10} $78 \pm 82$ \\ \hline
\multirow{2}{*}{\textit{BIBFA}}  & \textbf{0.89} & $0.87 \pm 0.004$\\
 & \cellcolor{gray!10} 200 & \cellcolor{gray!10} $180 \pm 10$ \\
\hline \hline
\multirow{2}{*}{CCA} &  0.88 & $0.87 \pm 0.001$ \\ 
 & \cellcolor{gray!10} 72 & \cellcolor{gray!10} 72 \\ \hline
\multirow{2}{*}{CCA + Log. Reg.} & \textbf{0.89} & $0.87 \pm 0.002$  \\ 
 & \cellcolor{gray!10} 72 & \cellcolor{gray!10} 72 \\ \hline
\multirow{2}{*}{PCA + Log. Reg.} &  0.81 & $0.82 \pm 0.004$ \\ 
 & \cellcolor{gray!10} 252 & \cellcolor{gray!10} $18 \pm 0$ \\ \hline
\multirow{2}{*}{MLP} & 0.85 & $0.77 \pm 0.007$ \\
 & \cellcolor{gray!10} 50 & \cellcolor{gray!10} $210 \pm 37$  \\ \hline
Logistic reg. & \textbf{0.89} & $\mathbf{0.88 \pm 0.001}$ \\ \hline
Ridge reg. & \textbf{0.89} & $0.87 \pm 0.002$ \\
\hline \hline
\end{tabular}
\end{adjustbox}
\caption{Results on \textit{AVIRIS} database of the predictive SSHIBA and the different methods under study. The table shows the results modelling the SSHIBA algorithm treating the labels as multilabel (\textit{SSHIBA - multilabel}) and as categorical (\textit{SSHIBA - categorical}). Results include the performance in terms of AUC (white cells) and the number of latent factors (grey cells) when the complete dataset is used and when only $20\%$ of the training samples are used. The results on the reduced dataset have been calculated 5-fold CV, so their standard deviations are also included.}
\label{tab:satelliteTotal}
\end{table}

In Table \ref{tab:satelliteTotal} we have the results with the \textit{AVIRIS} database, where in this case the model aims at predicting a categorical variable. The conclusions drawn are similar w.r.t. the \textit{yeast} database. SSHIBA, particularly in the low-sample size case, can perform as good as the other base lines providing extra capabilities, as we demonstrate in the rest of experiments (feature selection, missing data imputation, multi-view learning). Note that we also include the SSHIBA performance when we treat the target variable as categorical (the true data type) or a multi-dimensional binary variable using its one-hot encoding. As expected, the performance improves when the data type is treated accordingly to its nature. 


\subsection{Feature selection with SSHIBA}

This section focuses on the extension of our model to allow feature selection, as presented in Section \ref{sec:Sparse}. To do so, we use the categorical and multilabel databases LFW and LFW-A. With these experiments we aim to visually analyse the feature relevances, as well as the latent space learnt by the model and how it describes the data.

\begin{figure}[h!t]
  \centering
  \begin{subfigure}[t]{0.45\textwidth}
    \centering
    \includegraphics[width=\linewidth]{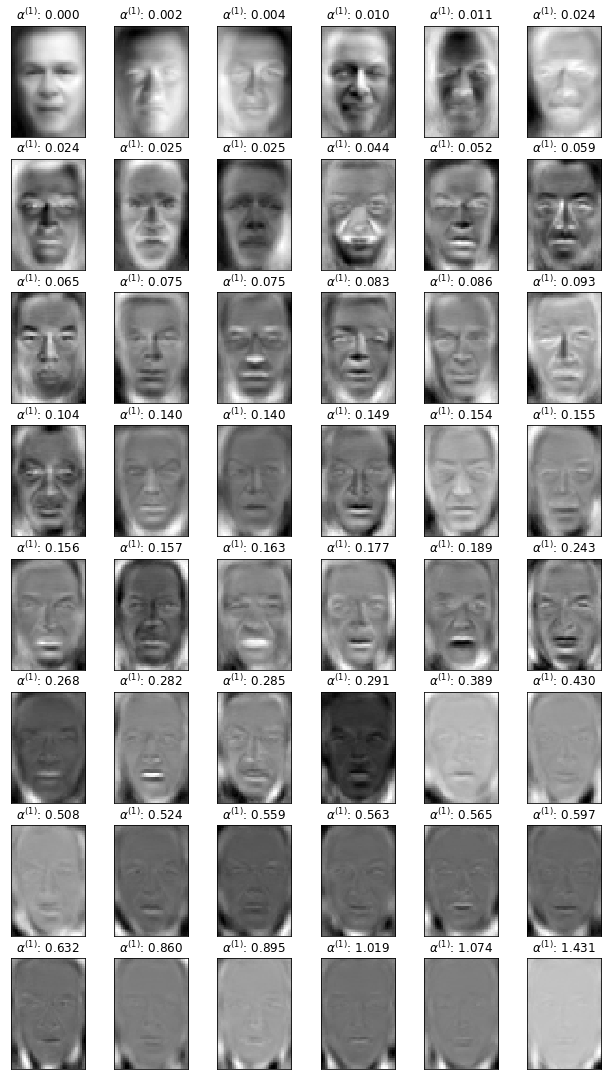}
    \caption{LFW database.}
    \label{fig:W_LFW}
  \end{subfigure}
  ~ 
  \begin{subfigure}[t]{0.45\textwidth}
    \centering
    \includegraphics[width=\linewidth]{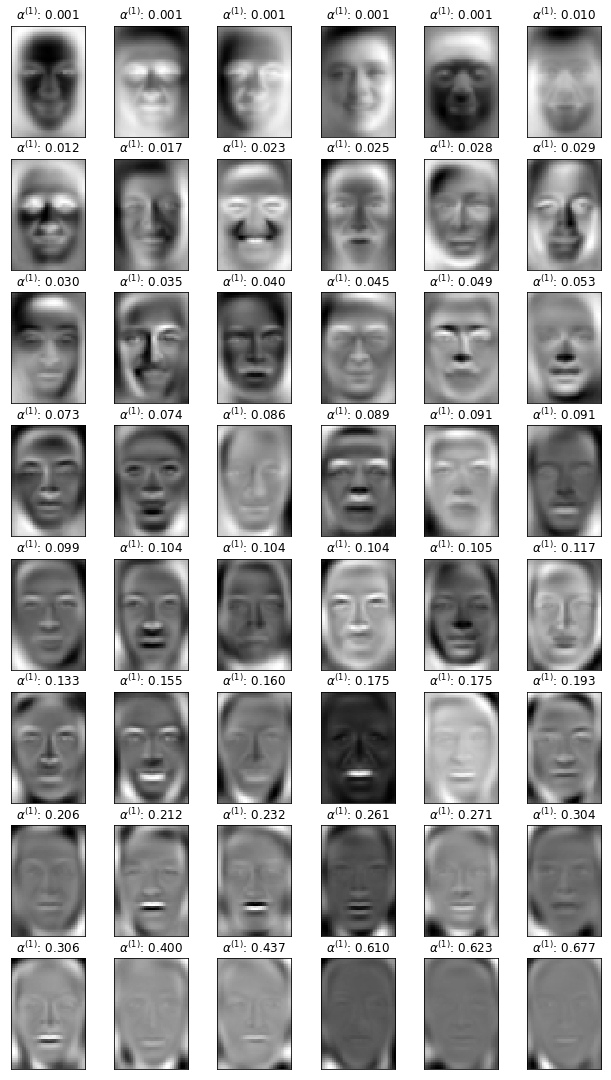}
    \caption{LFWA database.}
    \label{fig:W_LFWA}
  \end{subfigure}
  \caption{$\Wone$ matrix learnt by the sparse version of SSHIBA using two different databases. Each latent face is a column of this matrix $\Wone$. The images include the latent faces learned by the model, as well as the associated value of the variable $\aone$, which determines the relevance of each learnt column.}
  \label{fig:W}
\end{figure}

Figure \ref{fig:W} shows each of the columns of the matrix $\Wone$ learned by the model in both databases (recall that each column of this matrix has the same dimension as the images). The columns are ordered using the value of the variable $\aone$, since it provides the relevance of each latent factor. Note that each column of the matrix is capturing a face shape, and these faces will be combined for data reconstruction. In both Figure \ref{fig:W_LFW} and Figure \ref{fig:W_LFWA}, we can see how, as we advance through the faces, we reach a value of $\aone \approx 0.3$ in which the images become more blurry and less informative. It is around this point that we could start pruning and removing the irrelevant latent factors which do not provide significant information. 

Besides, these images reveal how the model adapts to the learning task. E.g., in the case of Figure \ref{fig:W_LFW} we can see how the model pay more attention to the different individuals and some latent-faces can be related to some labels: the first latent-face seems to be dedicated to \textit{George W. Bush} and the second one to \textit{Hugo Chavez}. On the other hand, in Figure \ref{fig:W_LFWA} latent faces tend to focus on face regions associated to different attributes, such as, the eyes area or the forehead.

\begin{figure}[h!t]
  \centering
  \begin{subfigure}[t]{0.38\textwidth}
    \centering
    \includegraphics[width=\linewidth]{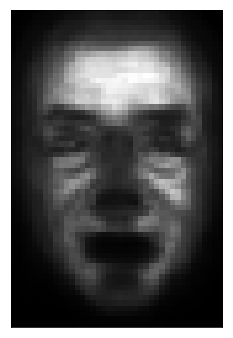}
    \caption{LFW database.}
    \label{fig:Gamma_LFW}
  \end{subfigure}
  ~ 
  \begin{subfigure}[t]{0.38\textwidth}
    \centering
    \includegraphics[width=\linewidth]{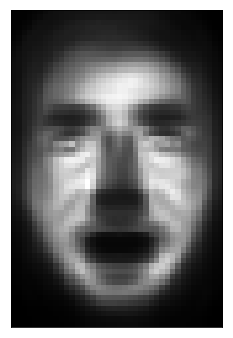}
    \caption{LFWA database.}
    \label{fig:Gamma_LFWA}
  \end{subfigure}
  \caption{Gamma masks learnt by the sparse version of SSHIBA using two different databases. The masks represent the importance of each pixel: lighter colours imply the pixel is more relevant while darker ones represent the pixel is less relevant.}
  \label{fig:Gamma}
\end{figure}

At this point we can visualise not only the projection matrices, but also the variable $\gamone$ which is the one used to include the sparsity in the input features, the pixels in our case. Figure \ref{fig:Gamma} provides this representation, which indicates which pixels are the most relevant for the problem and which are not. Again, we can see how the model adapts the results to the problem. For example, for the identification of 7 different subjects, Figure \ref{fig:Gamma_LFW} shows how the algorithm focuses on some specific areas, such as the forehead, to identify which of the subjects the image corresponds to. However, when looking at Figure \ref{fig:Gamma_LFWA} we can see that the model focuses on completely different regions. This is because in this case the model does not need to identify to whom the image corresponds, but to identify what characteristics the person has, such as his eye color, his race or the strength of his nose lines. 

\begin{figure}[h!t]
  \centering
  \begin{subfigure}[t]{0.48\textwidth}
    \centering
    \includegraphics[width=\linewidth]{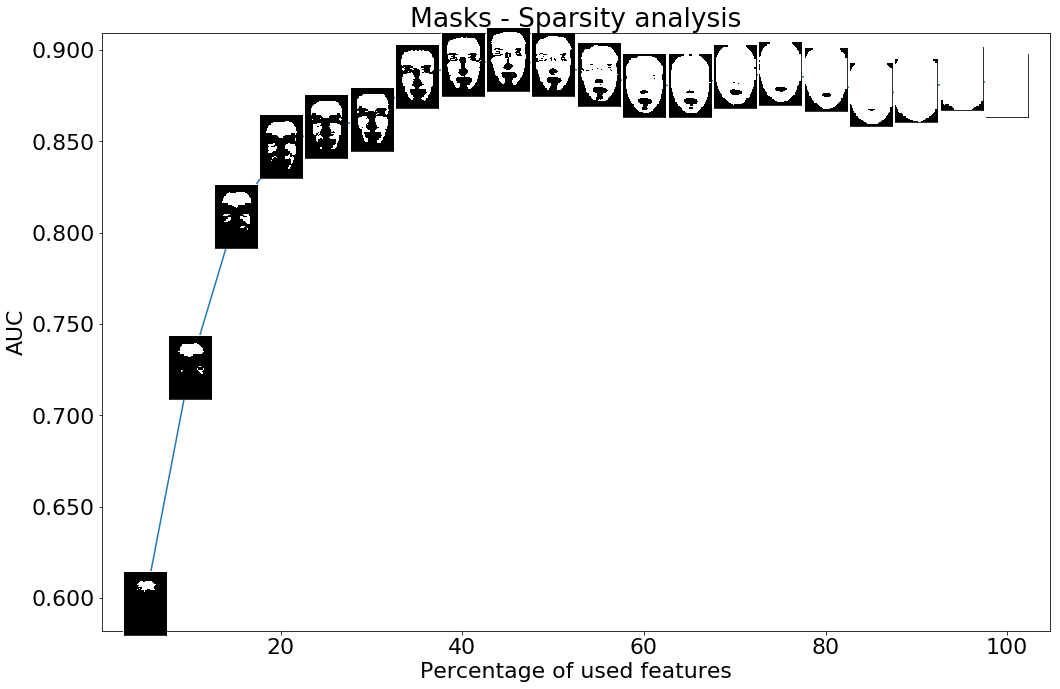}
    \caption{LFW database.}
    \label{fig:masksAUC_LFW}
  \end{subfigure}
  ~ 
  \begin{subfigure}[t]{0.48\textwidth}
    \centering
    \includegraphics[width=\linewidth]{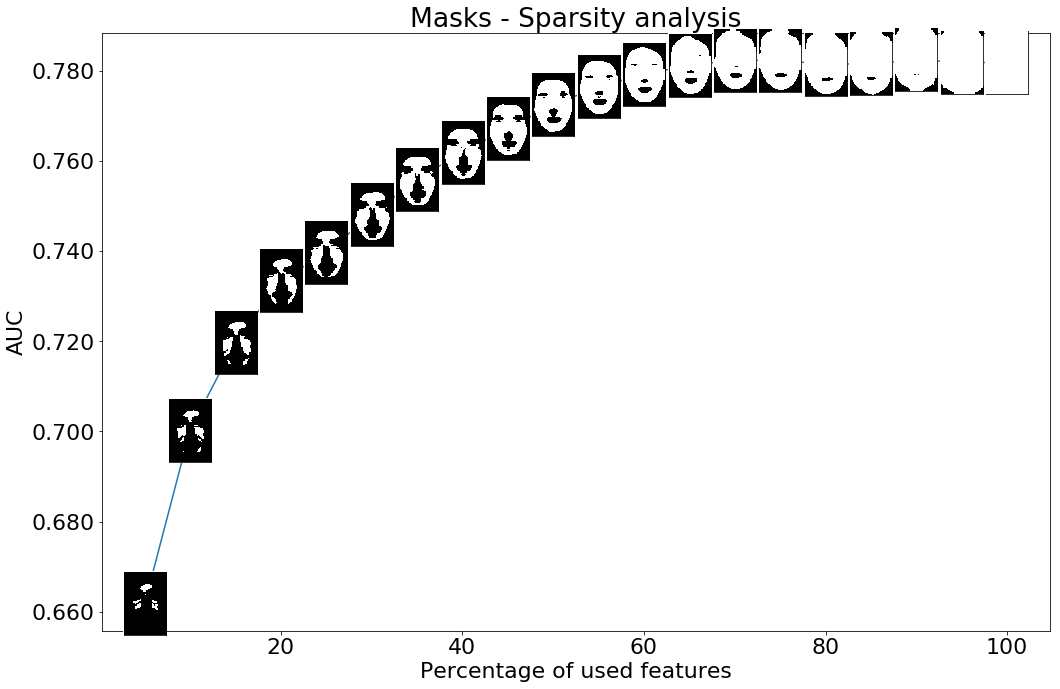}
    \caption{LFWA database.}
    \label{fig:masksAUC_LFWA}
  \end{subfigure}
  \caption{AUC results on the \textit{LFW} and \textit{LFWA} databases using the sparse version of the method. These images show the AUC results using different percentage of the most relevant values in the learnt mask. Each face shows the mask with different numbers of features.}
  \label{fig:masksAUC}
\end{figure}

Finally, we can analyze the goodness of this feature selection by calculating the final model performance for different percentages of selected features. For this purpose, we have ordered the features (pixels) by relevance and trained the model with different amounts of them. Figure \ref{fig:masksAUC} shows this AUC evolution. These results prove that using only around $50\%$ of the pixels, the model is capable of obtaining a good enough classification AUC. In particular, Figure \ref{fig:masksAUC_LFW} shows this result on the LFW database, in which the method does not need as many pixels to determine which subject the image belongs to, having a pretty good result using only $40 \%$ of the original pixels.


\subsection{Missing data imputation with SSHIBA}

This section presents the experiments we carried out using the semi-supervised approach for the imputation of missing values. In this case, we included random patterns of missing values in three different databases and used SSHIBA to impute such values using semi-supervised approach. For this experiment we used the \textit{yeast}, \textit{scene} and \textit{AVIRIS} databases. We compare the semi-supervised approach with both the predictive method (assuming no missing data in the train set), and with the results obtained when the train missing pattern is first imputed using some common imputation techniques. In Table \ref{tab:SS} we include the obtained results. First, note that in the case of no missing values, the semi-supervised method (which jointly processes the test and training data) is able to improve the predictive method, achieving a 0.68 AUC, which achieves the best result in Table \ref{tab:yeastTotal}. Furthermore, when we include a 50\% of missing data in the train set, the use of the semi-supervised SSHIBA with no pre-imputation method achieves the best results, as the probabilistic model is able to handle the uncertainty of the missing entries with no artificial imputation. This result certainly demonstrate the superior ability of the method to capture hidden correlations in our data, boosted by a proper modelling of each data type.

\begin{table}[hbt]
\centering
\begin{adjustbox}{max width=\textwidth}
\begin{tabular}{c|c|c|cccc}
\hline
\multirow{2}{*}{Missing Pattern} & \multirow{2}{*}{Imputation Method (train)} & \multirow{2}{*}{SSHIBA} & \multicolumn{3}{c}{AUCs}\\
&&& \textit{yeast} & \textit{AVIRIS} & \textit{scene} \\
\hline \hline
\multirow{2}{*}{No missing at train} & \multirow{2}{*}{--} & Predictive & 0.66 & 0.88 & 0.92 \\
 &  & SS & \textbf{0.68} & 0.88 & 0.92 \\\hline
\multirow{4}{*}{$50\%$ missing at train} & Semi-Supervised & \multirow{4}{*}{SS} & \textbf{0.64} & \textbf{0.87} & \textbf{0.89} \\
& Mean &  & 0.61 & 0.78 & 0.87\\ 
& Median &  & 0.55 & 0.78 & 0.70\\
& Most frequent value &  & 0.48 & 0.77 & 0.52\\ \hline \hline
\end{tabular}
\end{adjustbox}
\caption{Results on \textit{yeast}, \textit{scene} and \textit{AVIRIS} databases of the semi-supervised and predictive SSHIBA in comparison to different imputation techniques. Results include the AUC values when the complete dataset is used and when there is $50\%$ of missing input data.}
\label{tab:SS}
\end{table}


\subsection{Multiview learning with SSHIBA}

As a final experiment on the proposed SSHIBA algorithm, we tested its potential on a multiview problem. In this case, we decided to combine the information of the LFW database with the LFWA database, having information of both the person identity and the different characteristics that define it.  This problem was calculated with the previously defined baselines to compare the results. As these methods are not compatible with multiview, we decided to incorporate the extra information as an extra input feature.

\begin{table}[h!t]
\centering
\begin{adjustbox}{max width=\textwidth}
\begin{tabular}{r|cc}
\hline
 &  Two views & Three views \\\hline \hline
\multirow{2}{*}{\textit{SSHIBA}} & \textbf{0.68} & \textbf{0.69}\\
 & \cellcolor{gray!10} 39 & \cellcolor{gray!10} 35 \\
\hline \hline
\multirow{2}{*}{CCA} & 0.60 & 0.60\\ 
& \cellcolor{gray!10} 62 & \cellcolor{gray!10} 62\\\hline
\multirow{2}{*}{CCA + Log. Reg.} & 0.60 & 0.60\\ 
 & \cellcolor{gray!10} 62 & \cellcolor{gray!10} 62\\\hline
\multirow{2}{*}{PCA + Log. Reg.} & 0.65 & 0.66\\ 
 & \cellcolor{gray!10} 187 & \cellcolor{gray!10} 187 \\\hline
\multirow{2}{*}{MLP} & 0.60 & 0.60\\
 &\cellcolor{gray!10} 375 & \cellcolor{gray!10} 375\\\hline
Logistic reg. & 0.65 & 0.65\\ \hline
Ridge reg. & 0.67 & 0.67\\
\hline \hline
\end{tabular}
\end{adjustbox}
\caption{Results on \textit{LFWA} database using the data of the \textit{LFW} database as an extra view. Results include the performance in terms of AUC (white cells) and the number of latent factors (grey cells) when the complete \textit{lfwa } dataset is used (\textit{two views}) and when the data from the \textit{lfw} database is also used (\textit{three views}). The initial number of latent factors for both SSHIBA and BIBFA was 200. The results on the reduced dataset have been calculated 5-fold CV, so their standard deviations are also included.}
\label{tab:lfwaMultiviewMultilabel}
\end{table}

In Table \ref{tab:lfwaMultiviewMultilabel} we can see the results obtained. Results include the AUC values for all methods under study when the complete \textit{LFWA} dataset is used (\textit{Two views}) and when the data from the \textit{LFW} database is also incorporated (\textit{Three views}). First of all, we can notice that the SSHIBA algorithm is not only outperforming the rest of the baseline results, but also having a significantly lower number of latent features than the FE algorithms. Equivalently, we can see that the addition of a new view with further information on the data leads to a reduction on the latent features as well as an improvement of the performance of the algorithm. The inclusion of additional information allows the model to capture more accurate data correlations with a smaller hidden dimensionality.

\section{Conclusions}
\label{sec:conclusion}

In this article we generalize the BIBFA model to create a new FA framework, called SSHIBA, capable of adapting to the particularities of any learning problem. In particular, this new model includes new functionalities, such as, being able to carry out a selection of the most relevant features while extracting latent features, modelling not only real problem but also multilabel and categorical ones and, at the same time, work in a semi-supervised way with unlabelled data and missing values.

The results with SSHIBA show that, in the worst case, the performance of the method is similar to the state-of-the-art algorithms while being able to find a reduced latent space, having less extracted features than classical feature extraction methods. When  feature selection capabilities are included, the algorithm is able, during the feature reduction, to keep enough information to improve the interpretability of the results. Furthermore, when the data are modelled according to their nature (multilabel, categorical, ...) we obtain more compacted models (lower number of latent factors) maintaining or, in some cases, increasing the model performance.

On the other hand, the semi-supervised version of the algorithm has been proven to perform like the predictive or even outperform it, while providing the online imputation of any possible missing value in the data. One major advantage of our imputing method is that it is not only capable of working with some missing data in the input view but also is capable of imputing any kind of missing value in the labels. This advantage allows the algorithm to use a greater amount of datasets that might have some unclassified data. 

In short, the SSHIBA algorithm  is capable of doing feature selection, feature extraction and imputing the missing values at the same time while providing a good performance. Furthermore, the ability of working with multiple views combined with modelling the data according to their characteristics have provided more compact models together with a performance improvement.

\pagebreak

\appendix
\begin{changemargin}{-1cm}{-1cm}
\section*{\hfil \LARGE Appendices\hfil}
\section{Variational updates for SSHIBA with feature selection}
\label{sec:appendixA}

In the following we apply the mean-field update in (10) to the variational posterior family in (22).
\subsection*{\underline{$q(\Wm)$ update}}

\begin{align} 
&\lnp{q^*\p*{\Wm}} = \E_{\Z,\taum, \bim}\cor*{\lnp{p\p*{\Xm,\Wm,\Z,\am,\taum,\bim}}} \nonumber \\
&= \E_{\Z,\taum, \bim}\cor*{\lnp{p\p*{\Xm | \Wm,\Z,\taum,\bim}}} \nonumber \\
& + \E_{\am, \gamm}\cor*{\lnp{p\p*{\Wm | \am, \gamm}}} + \const \nonumber \\
\label{eq:lnqopt}
\end{align} 
We now evaluate both terms and then sum the results:
\begin{align} 
&\lnp{p\p*{\Xm | \Wm,\Z,\taum,\bim}} \nonumber \\
&= - \frac{\ang{\taum}}{2} \sumn \left(- 2 \ang{\Zn} \WmT \XnmT + \ang{\Zn \WmT \Wm \ZnT} + 2 \ang{\bim} \Wm \ang{\ZnT} \right) + \const \nonumber\\
&= - \frac{\taum}{2} \sumn\sumd \left(- 2 \ang{\Zn} \WdmT \XndmT + \ang{\Zn \WdmT \Wdm \ZnT} + 2 \ang{\bidm} \Wdm \ang{\ZnT} \right) + \const \nonumber\\
&= - \frac{\ang{\taum}}{2} \sumn\sumd \left(- 2 \XndmT \ang{\Zn} \WdmT + 2 \ang{\bidm} \ang{\Zn} \WdmT  + \Wdm \ang{\ZnT \Zn} \WdmT \right) + \const \nonumber\\
\label{eq:firstpW}
\end{align} 

and hence
\begin{align} 
&\E_{\Z,\taum, \bim}\cor*{\lnp{p\p*{\Xm | \Wm,\Z,\taum,\bim}}} \nonumber \\
\eqeq \sumn\sumd \left( \ang{\taum} \p*{\XndmT \ang{\Zn} - \ang{\bidm} \ang{\Zn} }\WdmT  \right. \nonumber \\
\eqline \left. - \frac{\ang{\taum}}{2} \Wdm \ang{\ZnT \Zn}\WdmT \right) + \const \nonumber\\
\label{eq:firstEW}
\end{align} 

The second term in \eqref{eq:lnqopt} can be calculated as:
\begin{align} 
\lnp{p\p*{\Wm | \am, \gamm}} 
\eqeq  \sumd \sumk \p*{ \frac{1}{2}\lnp{\akm \gamdm} - \frac{\akm}{2}{\Wdkm}^2}+ \const
\end{align}
and the expectation reads as follows
\begin{align} 
\E_{\am, \gamm}\cor*{\lnp{p\p*{\Wm | \am, \gamm}} } 
\eqeq \sumd \sumk \p*{ - \frac{1}{2} {\Wdkm}\ang{\akm}\ang{\gamdm}{\Wdkm}}+ \const 
\label{eq:secondEWSparse}
\end{align} 

If we put all together we get
\begin{align} 
& \lnp{q^*\p*{\Wm}} = \sumd \sumk \p*{ - \frac{1}{2} {\Wdkm}\ang{\akm}\ang{\gamdm}{\Wdkm}} + \sumn\sumd \left( \ang{\taum}   \right. \nonumber\\
& \left. \p*{\XndmT \ang{\Zn} - \ang{\bidm} \ang{\Zn} }\WdmT - \frac{\ang{\taum}}{2} \sumd \p*{\Wdm \ang{\ZnT \Zn}\WdmT}\right) + \const \nonumber \\
&= \sumd( - \frac{1}{2} \Wdm \p*{\text{diag}(\ang{\akm})\ang{\gamdm} + \ang{\taum}\ang{\ZnT \Zn}}\WdmT \nonumber\\
& + \ang{\taum} \sumn\p*{\XndmT \ang{\Zn} - \ang{\bidm} \ang{\Zn} }\WdmT) + \const,
\end{align}
from where we can identify that $q^*\p*{\Wm} $ follows a Gaussian distribution
\begin{align} 
q^*\p*{\Wm} \eqeq \prodd\p*{\N \p*{\Wdm | \mu_{\Wdm}, \Sigma_{W_d^{(m)}}}},
\label{eq:qWSparse}
\end{align} 
with covariance matrix
\begin{align} 
\Sigma_{\Wdm}^{-1} \eqeq diag(\ang{\am})\ang{\gamdm} + \ang{\taum}\ang{\ZT \Z}  
\label{eq:varWSparse}
\end{align} 
and mean
\begin{align} 
\mu_{\Wm} \eqeq \ang{\taum} \p*{\Xm - \mathbbm{1}_N \ang{\bim}}^T \ang{\Z} \Sigma_{\Wm}
\label{eq:meanWb}
\end{align} 
where $\mathbbm{1}_N$ is a row vector of ones of dimension $N$.

\subsection*{\underline{$q(\am)$ update}}

The update is summarized by the following expression
\begin{align}\label{eqalpha}
\lnp{q^*\p*{\am}} \eqeq \E_{\Wm}\cor*{\lnp{p\p*{\Wm | \am, \gamm}}} + \E\cor*{\lnp{p\p*{\am}}} + \const,
\end{align}
where
\begin{align} 
\lnp{p\p*{\Wm | \am, \gamm}} \eqeq \sumd\sumk \p*{ \frac{1}{2}\ln\p*{\akm \gamdm} - \frac{1}{2} \Wdkm\akm\gamdm\Wdkm} +\const \nonumber
\end{align}
and hence
\begin{align} 
&\E \cor*{\lnp{p\p*{\Wm | \am,\gamm}}} \nonumber \\
&=\sumk\left(\frac{D_m}{2}\lnp{\akm}- \frac{1}{2} \akm \sumd\p*{\ang{\gamdm}\ang{\WkmT \Wkm }}\right) + \const \label{eq:firstEAlphaSparse}
\end{align} 
Regarding the second term in \eqref{eqalpha} we have
\begin{align} 
\lnp{p\p*{\akm}} \eqeq - \beta_0 \akm + \p*{\alpha_0 -1}\lnp{\akm} + \const \nonumber
\end{align}
and the expectation is experessed as follows
\begin{align} 
\E \cor*{\lnp{p\p*{\am}}} 
\eqeq \sumk\p*{ \lnp{p\p*{\akm}} } =  \sumk\p*{ - \beta_0 \akm + \p*{\alpha_0 -1}\lnp{\akm}} + \const \label{eq:secondEAlpha}
\end{align} 
Combining both expectation we can conclude that
\begin{align} 
q^*\p*{\am} \eqeq \prodk \p*{ \text{Gamma}\p*{\akm | a_{\akm},b_{\akm}}}
\end{align} 
where
\begin{align} 
a_{\akm} \eqeq \frac{D_m}{2} + \alpha_0
\label{eq:aAlphaSparse}\\
b_{\akm} \eqeq \beta_0 + \frac{1}{2} \sumd\p*{\ang{\gamdm}\ang{\Wdkm\Wdkm}}
\label{eq:bAlphaSparse}
\end{align} 

\subsection*{\underline{$q(\gamm)$ update}}

The update closely follow the derivation of the $q(\am)$ update. From
\begin{align} 
\lnp{q^*\p*{\gamm}} \eqeq \E_{\Wm,\am}\cor*{\lnp{p\p*{\Wm | \am}}} + \E\cor*{\lnp{p\p*{\am}}} + \const
\end{align} 
we can easily obtain that
\begin{align} 
q^*\p*{\gamm} \eqeq \prodk \p*{ \text{Gamma}\p*{\gamdm | a_{\gamdm},b_{\gamdm}}}
\end{align} 
where
\begin{align} 
a_{\gamdm} \eqeq \frac{K_c}{2} + \alpha_0^\gamma
\label{eq:aGammaSparse}\\
b_{\gamdm} \eqeq \beta_0^\gamma + \frac{1}{2} \sumk\p*{\ang{\akm}\ang{\Wdkm\Wdkm}}
\label{eq:bGammaSparse}
\end{align} 
\end{changemargin}

\section{Multidimensional binary views}
\label{sec:appendixB}
\begin{changemargin}{-1cm}{-1cm}
The use of the lower-bound in (26) using the variational parameters $\bm{\xi}$ only affects the variational update for $\Xm$. To jointly optimize both $\bm{\xi}$ and $q(\Xm)$ we follow a variational expectation maximization (EM) procedure. First, for fixed $t^{(m)}$, by applying lower-bound in (26) along with the mean-field factorization we obtain (see \cite{jaakkola1997variational} for further details)
\begin{align} 
&\lnp{q^*\p*{\Xm}} = \E_{\Z,\Wm,\taum,\bim}\cor*{\lnp{p\p*{\Xm,\Wm,\Z,\am,\taum,\bim}}} \nonumber \\
&= \E\cor*{\lnp{h\p*{\Xtwo,\xi}}} + \E_{\Z,\Wm,\taum,\bim}\cor*{\lnp{p\p*{\Xm | \Wm,\Z,\taum,\bim}}} + \const \nonumber \\
\end{align}
where $h\p*{\Xm,\bm{\xi}}$ is defined in (26). A straightforward calculation shows that
\begin{align}
&\E\cor*{\lnp{h\p*{\Xm,\bm{\xi}}}} \\&= \E\cor*{\sumn\sumdm\p*{\lnp{\sigma\p*{\xi_{n,d}}}+\Xndm \tndm - \frac{1}{2}\p*{\Xndm + \xi_{n,d}}- \lambda\p*{\xi_{n,d}}\p*{\Xndm^2 - {\xi_{n,d}}^2}}} \nonumber\\
&= \sumn\sumdm\p*{\Xndm \tndm - \frac{1}{2}\Xndm - \lambda\p*{\xi_{n,d}}\Xndm^2} +\const \nonumber\\
&= \sumn\p*{\p*{\tnm  - \frac{1}{2}} \XnmT - \Xnm\Lambda_{\bm{\xi}_{n,:}}\XnmT} +\const\nonumber 
\end{align}
where $\Lambda_{\bm{\xi}_{n,:}}$ is a diagonal matrix which diagonal elements are $\lambda\p*{\xi_{n,1}}, \lambda\p*{\xi_{n,2}}, \dots, \lambda\p*{\xi_{n,D_2}}$. Also, we have that
\begin{align} 
&\lnp{p\p*{\Xm | \Wm,\Z,\taum}} = \sumn \lnp{\N\p*{ \Zn \WmT + \bim, \p*{\taum}^{-1}I}}+ \const \nonumber \\
\eqeq - \frac{\taum}{2} \sumn \p*{\Xnm\XnmT - 2 \p*{\Zn \WmT + \bim} \XnmT } + \const \nonumber
\label{eq:pXbias}
\end{align} 
Therefore, joining both terms we have that:
\begin{align}
\lnp{q^*\p*{\Xm}} \eqeq \sumn\left(\p*{\tnm - \frac{1}{2} + \ang{\taum} \p*{\ang{\Zn}\ang{\WmT}} + \ang{\bim}} \XnmT \right. \nonumber \\
\eqline \left.-\frac{1}{2}\Xnm\p*{\ang{\taum} I + 2\Lambda_{\xi_n}} \XnmT \right)+\const
\end{align}

This way we obtain that the distribution is as follows:
\begin{align}
q^*\p*{\Xm} \eqeq \prodn \p*{ \N \p*{\Xnm | \ang{\Xnm},\Sigma_{\Xm}}}
\end{align}
\begin{align}
\Sigma_{\Xnm}^{-1} \eqeq \ang{\taum} I_{D_m} + 2\Lambda_{\xi_n}
\end{align}
\begin{align}
\ang{\Xntwo} \eqeq \p*{\tnm - \frac{1}{2} + \ang{\taum} \p*{\ang{\Zn}\ang{\WmT} + \ang{\bim}}}\Sigma_{\Xnm}
\end{align}

\subsection{Variational parameter calculation (\texorpdfstring{$\xi_{n,d}$}{TEXT})}

The optimization of the variational parameters in $\bm{\xi}$ is done by equalizing the gradient of the ELBO lower  bound in (8) w.r.t. $\bm{\xi}$  to zero. By combining the expression for the ELBO lower bound with the conditional distribution lower-bound in (26), it can be shown that the only terms that depend on $\bm{\xi}$ are 
\begin{align}
 L = \E_{q}\cor*{\lnp{h\p*{\Xm,\bm{\xi}}}} + \E_{q}\cor*{\lnp{q\p*{\Xm}}},
\end{align}
where
\begin{align}
\E_{q}\cor*{\lnp{p\p*{\tm|\Xm}}} 
\eqeq \E_{q}\cor*{\lnp{h\p*{\Xm,\bm{\xi}}}} \nonumber\\
\eqeq \sumn\sumd \left(\lnp{\sigma\p*{\xi_{n,d}}} + \ang{\Xndm} \tndm - \frac{1}{2}\p*{\ang{\Xndm} + \xi_{n,d}} \right. \nonumber\\
\E_{q}\cor*{\lnp{q\p*{\Xm}}} 
\eqeq  \sumn\p*{\frac{D_m}{2} \lnp{2\pi e} + \frac{1}{2} \ln|\Sigma_{\Xnm}|} \label{eq:ElogpXtwo}
\end{align}

We can now set the derivative with respect to the parameter $\xi_{n,d}$ equal to zero:
\begin{align}
\frac{\partial L}{\partial \xi_{n,d}} \eqeq \lambda'\p*{\xi_{n,d}}\p*{\E\cor*{\Xndm^2} - {\xi_{n,d}}^2} = 0
\end{align}
where $\lambda\p*{\xi_{n,d}}$ is defined in (25) and is is a monotonic function of $\xi_{n,d}$ for $\xi_{n,d} \geq 0$. Hence, if we ignore the $\xi_{n,d}=0$ solution, we have
\begin{align}
\lambda'\p*{\xi_{n,d}} \neq 0 \longrightarrow {\xi_{n,d}^{new}}^2 = \E\cor*{\Xndm^2}  = \ang{\Xnm}\ang{\Xnm}^T + \Sigma_{\Xnm}
\end{align}
\end{changemargin}

\section{Semisupervised model}
\label{sec:appendixC}
\begin{changemargin}{-1cm}{-1cm}
Missing views are handled as any other random variable in the model and hence during variational inference our goal is now to approximate the joint posterior distribution of the parameters of the model $\Theta$ and the missing data views ($\XmS$ or  $\tmS$). In the following, we apply the mean-field update rule to the missing data factors in the variational family in \eqref{eq:qModel_SSapp}.

\begin{align}
p(\Theta, \mathbf {\tilde{T}}^{\{\mathcal{M}_t\}},\mathbf {\tilde{X}}^{\{\mathcal{M}_r\}} &| \tmt,\mathbf {X}^{\{\mathcal{M}_r\}}) \eqapprox \nonumber \\
& q\p*{\Z} \prod_{m_t \in \mathcal{M}_t} \left(q(\mathbf {\tilde{T}}^{(m_t)})\prod_{n=1}^{N}q\p*{\xnmmt}\right) \nonumber\\
&\times \prod_{m_r \in \mathcal{M}_r}q(\mathbf {\tilde{X}}^{(m_r)})\prodm\p*{q\p*{\Wm} q\p*{\am} q\p*{\taum} q\p*{\gamm}} \label{eq:qModel_SSapp}
\end{align}

\subsection{Unobserved real-valued views}

For any unobserved real-valued data view $\tilde{\mathbf{x}}^{(m)}_{n,:}$, the mean-field update equation can be simplified to 
\begin{align}
\lnp{q(\tilde{\mathbf{x}}^{(m)}_{n,:})}&=  \E\cor*{\lnp{p\p*{\tilde{\mathbf{x}}^{(m)}_{n,:} | \mathbf{W}^{(m)}, \Zn, \tau^{(m)}}}} + \const \nonumber \\
&=- \frac{\taum}{2} \p*{\tilde{\mathbf{x}}^{(m)}_{n,:}(\tilde{\mathbf{x}}^{(m)}_{n,:})^T - 2 \tilde{\mathbf{x}}^{(m)}_{n,:}\ang{\Wm} \ang{\Zn}^T }+ \const,
\end{align}
from which we identify that $q(\tilde{\mathbf{x}}^{(m)}_{n,:})$ corresponds to a Gaussian distribution with covariance matrix $ \ang{ \tau^{(m)} } \mathbf{I}$ and mean $\ang{\Zn} \WmT + \bim$. 

\subsection{Unobserved multidimensional binary view}

For unobserved multidimensional binary views, using the mean-field update can be reduced to
\begin{align} 
&\lnp{q\p*{\tmS}} = \E\cor*{\lnp{p\p*{\tmS | \XmS}}} = \E\cor*{\lnp{h\p*{\XmS, \xi}}} \nonumber \\
&= \E\cor*{\sum_{n=N^*}^N\sumdm\p*{\lnp{\sigma\p*{\xi_{nd}}}+\XndmS \tndmS - \frac{1}{2}\p*{\XndmS + \xi_{nd}}- \lambda\p*{\xi_{nd}}\p*{\XndmS^2 - {\xi_{nd}}^2}}} \nonumber \\
&=  \tmS \ang{\XmS}+ \const, 
\label{eq:logqtS}
\end{align} 
and therefore we have a logistic posterior distribution in which
\begin{align} 
q\p*{\tmS = 1} \eqeq \frac{e^{\ang{\XmS}}}{1+e^{\ang{\XmS}}} = \sigma\p*{\ang{\XmS}} \nonumber \\
q\p*{\tmS = 0} \eqeq  \frac{1}{1+e^{\ang{\XmS}}}
\end{align} 
where the denominator has been determined so that the sum of both probabilities is 1. The expected values is computed as
\begin{align} 
\ang{\tmS} \eqeq 1 * q\p*{\tmS = 1} + 0 * q\p*{\tmS = 0} = \sigma\p*{\ang{\XmS}},
\end{align} 
and it will be used in the update of the posterior distribution for $\Xm$. Namely, in Table 3 we replace $\tm$ by $\ang{\tmS}$ for those views in which $\tm$ is not observed.

\subsection{Unobserved categorical views}

Given an unobserved categorical variable $\tilde{t}_n^{(m)}$, note that the posterior factor $q(\tilde{t}_n^{(m)})$ correspond to a discrete categorical distribution with $D_m$ classes (the dimension of $\Xnm$). When $t_n^{(m)}=i$ is observed, the mean-field factor $q(\Xnm)$ in Table 4 correspond to a Gaussian distribution truncated at the space region in which $\delta\p*{\Xnimt > \Xnjmt \forall i\neq j}$. In the case of unobserved $\tilde{t}_n^{(m)}$, $q(\Xnm)$ will be defined by a mixture of truncated Gaussian distributions with weights given by $q(\tilde{t}_n^{(m)})$. Namely,
\begin{align}
q(\Xnm) &= \sum_{i=1}^{D_m} q(\Xnm)|_{\tilde{t}_n^{(m)}=i} ~~ q(\tilde{t}_n^{(m)}=i),\\
\ang{\Xnm} &= \sum_{i=1}^{D_m} <\Xnm>|_{\tilde{t}_n^{(m)}=i} ~~q(\tilde{t}_n^{(m)}=i),
\end{align}
where both $q(\Xnm)|_{\tilde{t}_n^{(m)}=i}$ and $<\Xnm>|_{\tilde{t}_n^{(m)}=i}$ are given in Table 4.

Finally, the mean-field update for $q(\tilde{t}_n^{(m)}=i)$
\begin{align}
    \log q(\tilde{t}_n^{(m)}=i) =\E [\log p(\tilde{t}_n^{(m)}=i|\Zn,\Wm)]
\end{align}
does not have analytic solution. It can be approximated by Monte Carlo (by iteratively sampling from $\Z$ and $\Wm$ from the posterior varitional mean field factors). However, the update using the mean values provides good results. Namely,
\begin{align}
    \log q(\tilde{t}_n^{(m)}=i) = \log p(\tilde{t}_n^{(m)}=i|\ang{\Zn},\ang{\Wm}).
\end{align}
\end{changemargin}

\pagebreak

\section{Acknowledgments}
\label{sec:acknowledgments}
The authors wish to thank Irene Santos, for fruitful discussions and help during the earlier stages of our work. The work of Pablo M. Olmos is supported by   Spanish  government   MEC   under   grant   PID2019-108539RB-C22,  by  Comunidad de   Madrid   under   grants   IND2017/TIC-7618, IND2018/TIC-9649, and Y2018/TCS-4705,  by BBVA Foundation under the Deep-DARWiN project,  and   by   the European  Union  (FEDER and the European Research Council (ERC) through the European Unions Horizon 2020 research and innovation program under Grant 714161). C. Sevilla-Salcedo and V. G{\'o}mez-Verdejo's work has been partly funded by the Spanish MINECO grants TEC2014-52289R and TEC2017-83838-R.

\label{sec:references}
\bibliographystyle{ieeetr}
\bibliography{bibliography}

\end{document}

%% file: article.definitions.tex
\newcommand{\E}{\mathbb{E}} 
\newcommand{\N}{\mathcal{N}} 
\newcommand{\M}{\mathcal{M}} 
\newcommand{\const}{\text{const}} 
\DeclareMathOperator{\Tr}{Tr} 

\DeclarePairedDelimiter\p{(}{)}
\DeclarePairedDelimiter\cor{[}{]}
\DeclarePairedDelimiter\llav{\{}{\}}
\DeclarePairedDelimiter\ang{\langle}{\rangle}

\DeclareMathOperator{\Xm}{\mathbf{X}^{(m)}} 
\DeclareMathOperator{\Xmmt}{\mathbf{X}^{(m_t)}} 
\DeclareMathOperator{\xnmmt}{\mathbf{x}^{(m_t)}_{n,:}} 
\DeclareMathOperator{\XM}{\mathbf{X}^{\{\M\}}} 
\DeclareMathOperator{\Xtwo}{\mathbf{X}^{(2)}} 
\DeclareMathOperator{\XmT}{\mathbf{X}^{(m)^T}} 
\DeclareMathOperator{\Xnm}{\mathbf{x}_{n,:}^{(m)}} 
\DeclareMathOperator{\XnMlast}{\mathbf{x}_{n,:}^{(M)}} 
\DeclareMathOperator{\XnM}{\mathbf{x}_{n,:}^{\{\M\}}} 
\DeclareMathOperator{\Xnone}{\mathbf{x}_{n,:}^{(1)}} 
\DeclareMathOperator{\Xntwo}{\mathbf{x}_{n,:}^{(2)}} 
\DeclareMathOperator{\XnmT}{\mathbf{x}_{n,:}^{(m)^T}} 
\DeclareMathOperator{\Xndm}{x_{n,d}^{(m)}} 
\DeclareMathOperator{\Xnim}{x_{n,i}^{(m)}} 
\DeclareMathOperator{\Xnimt}{x_{n,i}^{(m_t)}} 
\DeclareMathOperator{\Xnjmt}{x_{n,j}^{(m_t)}} 
\DeclareMathOperator{\XndmT}{x_{n,d}^{(m)^T}} 
\DeclareMathOperator{\xndmtwo}{x_{n,d}^{(m)^2}}

\DeclareMathOperator{\XmS}{\mathbf{\tilde{X}}^{(m)}} 
\DeclareMathOperator{\XnmS}{\mathbf{x}_{\ast,:}^{(m)}} 
\DeclareMathOperator{\XndmS}{\tilde{x}_{n,d}^{(m)}} 

\DeclareMathOperator{\xmS}{\mathbf{x}^{(m)}_{\ast,:}} 
\DeclareMathOperator{\xminS}{\mathbf{x}^{\{\M_{in}\}}_{\ast,:}} 
\DeclareMathOperator{\xmoutS}{\mathbf{x}^{\{\M_{out}\}}_{\ast,:}} 

\DeclareMathOperator{\ynm}{\mathbf{y}_{n,:}^{(m)}}
\DeclareMathOperator{\ynmt}{\mathbf{y}_{n,:}^{(m_t)}}
\DeclareMathOperator{\ynim}{\mathbf{y}_{n,i}^{(m)}}
\DeclareMathOperator{\ynimt}{\mathbf{y}_{n,i}^{(m_t)}}
\DeclareMathOperator{\ynjm}{\mathbf{y}_{n,j}^{(m)}}
\DeclareMathOperator{\ynjmt}{\mathbf{y}_{n,j}^{(m_t)}}
\DeclareMathOperator{\ynkmt}{\mathbf{y}_{n,k}^{(m_t)}}

\DeclareMathOperator{\ySnmt}{\mathbf{\tilde{y}}_{n,:}^{(m_t)}}

\DeclareMathOperator{\ySnimt}{\mathbf{\tilde{y}}_{n,i}^{(m_t)}}

\DeclareMathOperator{\ySnjmt}{\mathbf{\tilde{y}}_{n,j}^{(m_t)}}
\DeclareMathOperator{\ySnkmt}{\mathbf{\tilde{y}}_{n,k}^{(m_t)}}

\DeclareMathOperator{\Z}{\mathbf{Z}} 
\DeclareMathOperator{\ZS}{\mathbf{\tilde{Z}}} 
\DeclareMathOperator{\ZT}{\mathbf{Z}^T} 
\DeclareMathOperator{\Zn}{\mathbf{z}_{n,:}} 
\DeclareMathOperator{\ZnS}{\mathbf{\tilde{z}}_{n,:}} 
\DeclareMathOperator{\ZnT}{\mathbf{z}_{n,:}^T} 

\DeclareMathOperator{\zS}{\mathbf{z}_{\ast,:}} 

\DeclareMathOperator{\Wm}{\mathbf{W}^{(m)}} 
\DeclareMathOperator{\Wmout}{\mathbf{W}^{\{\M_{out}\}}} 
\DeclareMathOperator{\Wone}{\mathbf{W}^{(1)}} 
\DeclareMathOperator{\WmT}{\mathbf{W}^{(m)^T}} 
\DeclareMathOperator{\WmmtT}{\mathbf{W}^{(m_t)^T}} 
\DeclareMathOperator{\WmoutT}{\mathbf{W}^{\{\M_{out}\}^T}} 
\DeclareMathOperator{\Wkm}{\mathbf{w}_{:,k}^{(m)}} 
\DeclareMathOperator{\WkmT}{\mathbf{w}_{:,k}^{(m)^T}} 
\DeclareMathOperator{\Wdm}{\mathbf{w}_{d,:}^{(m)}} 
\DeclareMathOperator{\WdmT}{\mathbf{w}_{d,:}^{(m)^T}} 
\DeclareMathOperator{\Wdkm}{w_{d,k}^{(m)}} 

\DeclareMathOperator{\bim}{\text{\boldmath$b$}^{(m)}} 
\DeclareMathOperator{\bidm}{\text{$b$}_d^{(m)}} 
\DeclareMathOperator{\bimT}{\text{\boldmath$b$}^{(m)^T}} 

\DeclareMathOperator{\am}{\text{\boldmath$\alpha$}^{(m)}} 
\DeclareMathOperator{\aone}{\text{\boldmath$\alpha$}^{(1)}} 
\DeclareMathOperator{\akm}{\alpha_k^{(m)}} 

\DeclareMathOperator{\gamm}{\text{\boldmath$\gamma$}^{(m)}} 
\DeclareMathOperator{\gamone}{\text{\boldmath$\gamma$}^{(1)}} 
\DeclareMathOperator{\gamdm}{\gamma_d^{(m)}} 

\DeclareMathOperator{\taum}{\tau^{(m)}} 
\DeclareMathOperator{\taummt}{\tau^{(m_t)}} 
\DeclareMathOperator{\taumout}{\tau^{\{\M_{out}\}}} 

\DeclareMathOperator{\tm}{\mathbf{T}^{(m)}} 
\DeclareMathOperator{\tnm}{\mathbf{t}_{n,:}^{(m)}} 
\DeclareMathOperator{\tndotmmt}{t_{n,:}^{(m_t)}} 
\DeclareMathOperator{\tmS}{\mathbf{\tilde{T}}^{(m)}} 
\DeclareMathOperator{\tmt}{\mathbf{T}^{\{\M_t\}}} 
\DeclareMathOperator{\tnmS}{\mathbf{t}_{n,:}^{(m)*}} 
\DeclareMathOperator{\tndmS}{t_{n,d}^{(m)*}} 
\DeclareMathOperator{\tndm}{t_{n,d}^{(m)}} 

\DeclareMathOperator{\ttmS}{\mathbf{t}^{(m)*}} 
\DeclareMathOperator{\ttnm}{t_n^{(m)}} 
\DeclareMathOperator{\ttnmS}{t_n^{(m)*  }} 

\DeclareMathOperator{\sumd}{\sum\limits_{d=1}^{D_m}} 
\DeclareMathOperator{\sumdm}{\sum\limits_{d=1}^{D_m}} 
\DeclareMathOperator{\sumn}{\sum\limits_{n=1}^{N}} 
\DeclareMathOperator{\summ}{\sum\limits_{m=1}^{M}} 
\DeclareMathOperator{\sumk}{\sum\limits_{k=1}^{K_c}} 

\DeclareMathOperator{\prodd}{\prod\limits_{d=1}^{D_m}} 
\DeclareMathOperator{\prodn}{\prod\limits_{n=1}^{N}} 
\DeclareMathOperator{\prodk}{\prod\limits_{k=1}^{K_c}} 
\DeclareMathOperator{\prodm}{\prod\limits_{m=1}^{M}} 

\newcommand{\lnp}[1]{\ln\p*{#1}} 
\newcommand{\eqline}{\enskip&\enskip}
\newcommand{\eqeq}{\enskip=&\enskip}
\newcommand{\eqapprox}{\enskip\approx&\enskip}
\newcommand{\eqsimil}{\enskip\sim &\enskip}